\pdfoutput=1

\documentclass[11pt]{article}

\usepackage{acl}

\usepackage{times}
\usepackage{latexsym}
\usepackage{graphicx}
\usepackage{algorithm}
\usepackage{algpseudocode}
\usepackage{xspace}
\usepackage{multirow}
\usepackage{pifont}
\usepackage{xcolor}
\usepackage{booktabs}
\usepackage{color} 
\usepackage{diagbox}
\usepackage{rotating}
\usepackage{tabularx}

\usepackage{graphicx}
\usepackage{csquotes}
\usepackage{siunitx}
\usepackage{caption}
\usepackage{stfloats}
\usepackage{amsmath}
\usepackage{amsfonts}
\usepackage{bbm}
\usepackage{makecell}
\usepackage{multicol}

\usepackage[T1]{fontenc}

\usepackage[utf8]{inputenc}

\usepackage{microtype}

\newcommand{\benchmark}{TABLET\xspace}
\newcommand{\numucitasks}{10}
\newcommand{\numddxtasks}{10}
\newcommand{\numtotaltasks}{20}

\newcommand{\task}{t}

\newcommand{\I}{I}
\newcommand{\D}{D}
\newcommand{\x}{x} 
\newcommand{\y}{y}
\newcommand{\n}{n}
\newcommand{\F}{F}
\newcommand{\C}{C}
\newcommand{\f}{f}
\newcommand{\tr}{\textrm{train}}
\newcommand{\te}{\textrm{test}}

\newif\ifcomments
\commentstrue
\ifcomments
    \providecommand{\sameer}[2][]{{\protect\color{violet}{[\textbf{sameer}:\textbf{#1} #2]}}}
    \providecommand{\dylan}[2][]{{\protect\color{purple}{[\textbf{dylan}:\textbf{#1} #2]}}}
\else
    \providecommand{\sameer}[2][]{}
    \providecommand{\dylan}[2][]{}
\fi

\title{\benchmark: Learning From Instructions For Tabular Data}

\author{Dylan Slack \\
  UC Irvine \\
  \texttt{dslack@uci.edu} \\\And
  Sameer Singh \\
  UC Irvine \\
  \texttt{sameer@uci.edu}}

\begin{document}
\maketitle
\begin{abstract}
Acquiring high-quality data is often a significant challenge in training machine learning (ML) models for tabular prediction, particularly in privacy-sensitive and costly domains like medicine and finance.
Providing \textit{natural language instructions} to large language models (LLMs) offers an alternative solution.
However, it is unclear how effectively instructions leverage the knowledge in LLMs for solving tabular prediction problems.
To address this gap, we introduce \textit{\benchmark}, a benchmark of $\numtotaltasks$ diverse tabular datasets annotated with instructions that vary in their phrasing, granularity, and technicality.
Additionally, \benchmark includes the instructions' logic and structured modifications to the instructions.
We find in-context instructions increase zero-shot F1 performance for Flan-T5 11b by $44\%$ on average and $13\%$ for ChatGPT on \benchmark.
Also, we explore the limitations of using LLMs for tabular prediction in our benchmark by evaluating instruction faithfulness.
We find LLMs often ignore instructions and fail to predict specific instances correctly, even with examples.
Our analysis on \benchmark shows that, while instructions help LLM performance, learning from instructions for tabular data requires new capabilities.\footnote{Please find the \benchmark demo and code at \href{https://dylanslacks.website/Tablet}{\texttt{https://dylanslacks.website/Tablet}}.}
\end{abstract}

\section{Introduction}
\label{sec:intro}
Machine learning with tabular data is crucial across critical domains like health care and finance.
However, acquiring labeled data to train supervised learning models is often challenging.
For example, privacy restrictions in finance and medicine prevent data sharing~\cite{sharing-is-hard}.
Moreover, collecting data is very costly, imposing a significant financial barrier. 
These concerns make gathering training data challenging and sometimes impossible~\cite{Murdoch2021, bigdataconsumerprivacymoothy, crawford2014big, Cath2018-aj}.

Large language models (LLMs) have the potential to considerably lighten the burden of collecting data for tabular prediction.
 Trained on web-scale data, LLMs have extensive world knowledge that is useful for downstream tasks~\cite{NEURIPS2020_1457c0d6}.
Instead of gathering large datasets, practitioners could create models for tabular prediction by interactively providing \textit{natural language instructions} to LLMs.
For instance, a doctor knowledgeable about a rare disease could instruct an LLM to identify its symptoms. 
The model could then help identify the disease without sharing private training data or requiring data collection.
Since using instructions for tabular prediction with LLMs has considerable potential, we need to characterize both the benefits and limitations of such an approach.

\begin{figure*}
    \centering
    \includegraphics[width=\linewidth, trim=1.5cm 6.8cm 5.1cm 0.0cm, clip]{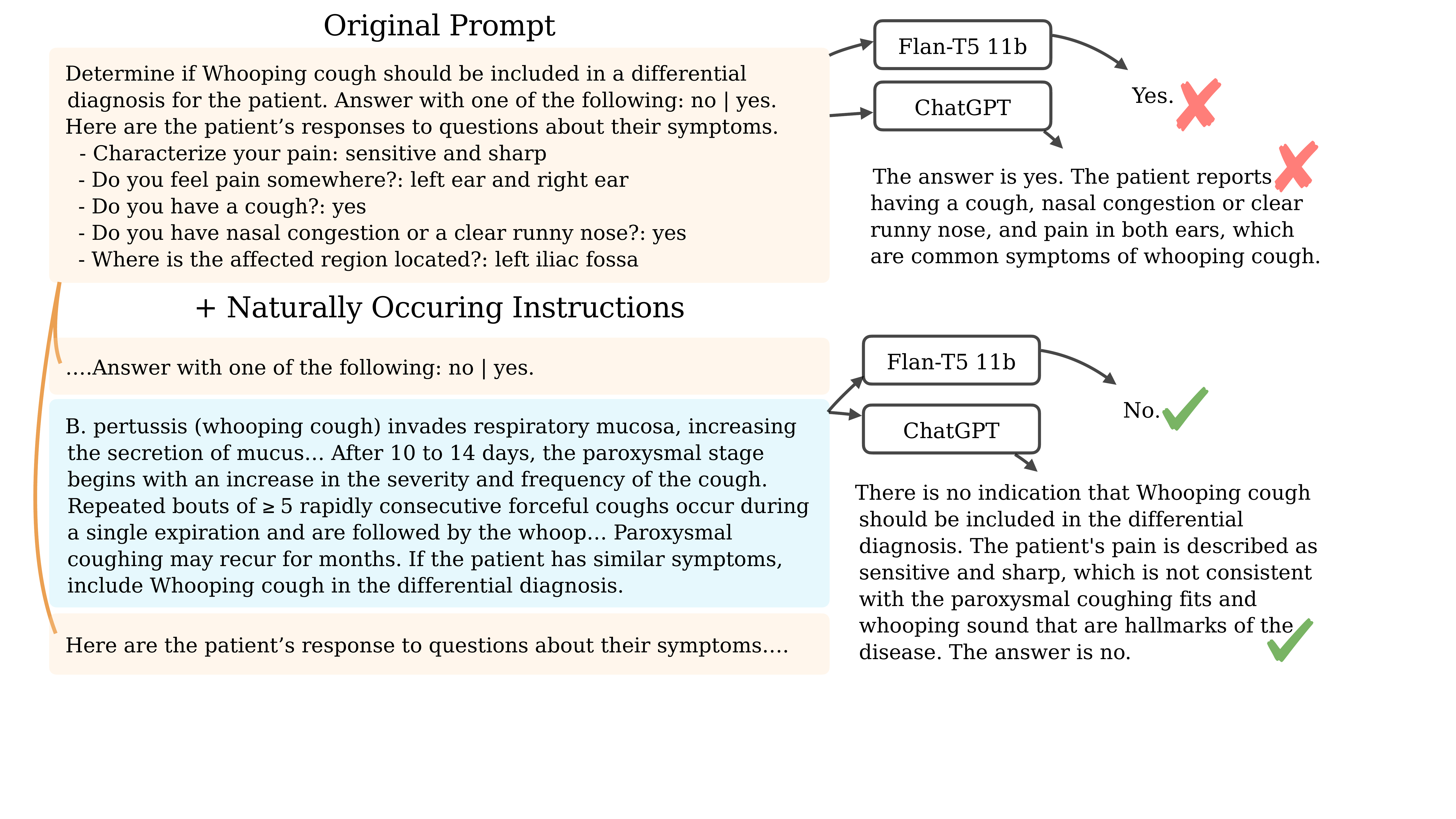}
    \caption{\textbf{Evaluation with \benchmark.} We serialize a short task description and a data point from the Whooping Cough dataset into a prompt (top). Flan-T5 11b and ChatGPT predict this instance incorrectly based on clues related to a cough and nasal congestion. However, by adding the instructions (bottom), the model identifies that the patient's pain is inconsistent with typical symptoms of Whooping Cough and correctly predicts the instance. 
    }
    \label{fig:fig1}
\end{figure*}

To evaluate the performance of LLMs at learning from instructions for tabular prediction, we introduce \textit{TAbular Benchmark for LEarning from Task instructions} (\benchmark).
This benchmark consists of $\numtotaltasks$ diverse tabular prediction tasks.
These include $\numucitasks$ datasets from the UCI ML repository, such as credit risk and churn prediction~\cite{Dua:2019}.
The additional $\numddxtasks$ tasks are differential diagnosis (DDX) datasets for diseases.
To robustly measure how effectively LLMs utilize instructions, we annotate the tasks with several generated and naturally occurring instructions categorized by their origin, structure, and level of technicality.

Using the instructions in \benchmark, researchers can evaluate how well models perform at tabular prediction by learning \textit{only} from in-context instructions (the \textit{zero-shot} setting) or in the \textit{few-shot} setting, where there are some labeled instances.
Moreover, they can contrast performance between collection sources, e.g., consumer versus technical professional references, or evaluate how to best structure the prompts.
We provide an example of evaluating with \benchmark in Figure~\ref{fig:fig1}.
When we only provide LLMs with the original data points serialized as text and a task description, LLMs fail to predict correctly.
However, when we provide explicit instructions for identifying symptoms of Whooping Cough in \benchmark, models consider more important features, such as paroxysmal coughing, and predict correctly.
Our analysis on \benchmark reveals zero-shot Flan-T5 11b improves F1 score by $20\%$ on average over the same model without instructions, and ChatGPT F1 performance improves $10\%$.
The performance gains increase in the few-shot setting; Flan-T5 11b improves on average $44\%$ over the baseline with $4$ in-context training examples, demonstrating instructions significantly improve LLM performance on tabular prediction.

We also evaluate the limitations and failure modes of learning from instructions for tabular prediction when using large language models.
We include additional artifacts in \benchmark, such as representations of the instruction's logic and \textit{flipped} instructions (i.e., instructions that, if followed, should lead to different answers than the originals).
In this setting, researchers can measure if models are faithful to instructions and use instructions for  generalization benefits beyond just following the information in the instructions.
While instructions improve LLM performance in general, we find LLMs are highly-biased against classifying certain instances correctly.
Similarly, LLMs do not always follow instructions provided in-context.
Overall, our analysis on \benchmark demonstrates instructions are promising for improving LLM performance on tabular prediction tasks, but current models have several key limitations.
In the future, we hope the release of \benchmark enables researchers to develop models capable of solving tabular prediction tasks from instructions alone.

\section{Tabular Instruction Learning}
\label{sec:method}
In this section, we introduce the problem of learning from instructions for tabular prediction.

\subsection{Problem Formulation}

This subsection formalizes our problem setting.

\paragraph{Tasks} Each tabular prediction task $t$ has dataset $\D_t = \{ (\x_i, \y_i ) \}_{i}^{\n}$, where data points $\x_i$ have $d$ features, the dataset $\D_t$ has $\n$ rows, and labels $\y_i$ belong to $w$ classes $\y_i \in \C_t$.
Additionally, the features and classes have names $\F_t=\{\f_1,...,\f_d\}$ and $\C_t = \{ c_1 ,..., c_w \}$, respectively.
We expect the feature names to be semantically meaningful, e.g., \textit{education} or \textit{marital status}, and the feature values to be in their normal range.
Last, each task has a title $E_t$ describing the goal (example in Table~\ref{table:income_prediction}).
We can often extract the feature names, class names, and titles from the metadata~\cite{Dua:2019}.

Also, the dataset $\D_t$ splits into training and testing sets, $\D_{t}^{\tr}$ and $\D_{t}^{\te}$ respectively.
While the focus of \benchmark is learning from instructions when there are \textit{few} or \textit{no} training examples in the training data $\D_{t}^{\tr}$, we provide the full training sets in \benchmark to compare against fully supervised models and perform few-shot evaluations.

\begin{table}[t]
\centering
\begin{tabular}{ll}
\toprule
 $E_t$ & Predict if income exceeds  \$50K/yr. \\
 $I_t$ & Individuals who earn more than \$50K/yr \\ & tend to have higher education levels \\ & (e.g., Bachelors or Prof-school). \\
$C_t$ & >50K | $\leq$50K \\
 $F_t$ & hours/week: 40 | sex:  Female | age: 24 \\& occupation:  Sales | education: college \\
\bottomrule
\end{tabular}
\caption{\textbf{Example Adult instance} (abbreviated), with title $E_t$, instructions $I_t$, classes $C_t$, and features $F_t$.}
\label{table:income_prediction}
\end{table}

\paragraph{Instructions} Each task $\task$ has \textit{natural language instructions} $\I_t$ that describe the relationship between the data $\D_t$ and labels $\C_t$.
The instructions may vary across many factors, such as style, granularity, and phrasing.
The goal is to use the instructions to solve the task with few or no training examples.

\subsection{Prompting Schema}
\label{subsec:schema}

We follow a consistent schema for collating tabular instruction learning problems into prompts, inspired by instruction learning for NLP~\cite{mishra-etal-2022-cross, supernaturalinstructions}.
For a task $t$, the prompts consist of five parts:
\begin{itemize}
    \item \textbf{Title:} The description of the task $E_t$.
    \item \textbf{Instructions:} The instructions $\I_t$.
    \item \textbf{Classes:} The classes $\C_t$ (e.g., ``yes'' or ``no'').
    \item \textbf{Examples:} In-context examples $\D_{t}^{\tr}$, if any.
    \item \textbf{Test Data Point:} Test example, $\x_i \in \D_{t}^{\te}$.
\end{itemize}
To convert tabular data points into strings for the \textit{Examples} and \textit{Test Data Point} components, we follow \citet{tabllm}.
We format feature names and values into strings: \texttt{``\{f$_1$\}:\{x$_1$\}\textbackslash n...\textbackslash n\{f$_d$\}:\{x$_d$\}''}.
We format the labels $\C_t$ into a string: \texttt{``Answer with one of the following:$\,$\{c$_1$\}|...|\{c$_w$\}''}.

\section{\benchmark}
\label{sec:tablet}

In this section, we introduce \benchmark.
The benchmark contains $\numtotaltasks$ tasks annotated with the components of the prompting schema (Subsection \ref{subsec:schema}).

\subsection{Tasks}

Here, we describe the tasks in \benchmark.

\paragraph{Differential Diagnosis} We include $10$ differential diagnosis (DDX) tasks derived from DDXPlus, a dataset for differential diagnosis carefully validated by doctors~\cite{tchango2022ddxplus}.
These tasks aim to predict if a disease, such as Ebola or Myocarditis, could cause a patient's symptoms (example in Fig.~\ref{fig:fig1}).
While the DDX problem is highly important, model development is limited by data privacy, making it an ideal task for instruction learning~\cite{Murdoch2021, NEURIPS2018_b5a1d925}.
These tasks have sparse binary features, indicating whether a patient has a specific symptom.
Because there are many features, we use only the positive features that indicate symptom presence in the prompts.
\begin{figure*}[ht]
    \centering
    \includegraphics[width=.9\linewidth, trim=0 19cm 0 0, clip]{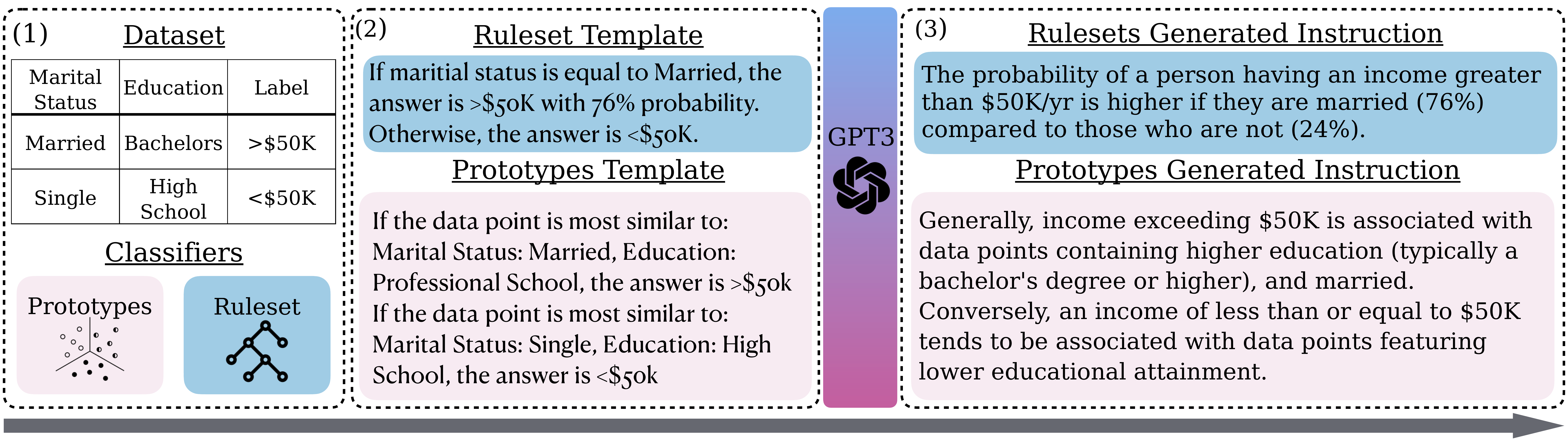}
    \caption{\textbf{Instruction generation pipeline.} (1) We fit a prototypes and rulesets classifier on the dataset; here Adult (2) We serialize the classifier's logic into a template (3) We sample revisions of the template with GPT-3.}
    \label{fig:annotation-overview}
\end{figure*}
\paragraph{UCI} We add $\numucitasks$ diverse tasks from the UCI ML Repository, such as  Customer Churn and Credit~\cite{Dua:2019}. 
These tasks represent many application areas and differ along dataset characteristics.
For instance, this set includes datasets with numeric, categorical, and mixed features and different class sizes (binary to $16$-way).

\paragraph{Splits} We split the datasets with an $80/20$ train/test split.
While we include the full training data in \benchmark, critically, our evaluation with LLMs takes place in the \textit{zero} and \textit{few-shot} settings. We include the full training set to permit comparison against fully supervised models and different few-shot samples.
Because the DDX tasks have many records (>$1$M), we down-sample to $10,000$ instances.
The DDX data is highly imbalanced, so we rebalance the data before splitting and sample the test set in proportion to the original distribution.

\subsection{Instructions}

This section describes the instruction annotations.

\subsubsection{Naturally Occurring Instructions}
We collect $3$ naturally occuring instruction annotations for each DDX dataset.
Two of these instructions are from consumer friendly sources, while one is from a more technical professional reference.
The consumer instructions are from government health websites and non-profits.
To ensure quality, we use articles in an index of consumer medical information maintained by the National Library of Medicine called MedlinePlus~\cite{medlineplus}.
We collect the more technical instruction from a professional reference called the Merck Manual~\cite{Bullers2016}.
To collect the instructions, first, we search for the disease using its ICD-10 code in either MedlinePlus or Merck.
ICD-10 is a standardized disease classification system that helps us ensure we find the correct information~\cite{WHO1993}.
Next, we find the symptoms section on the webpage.
We create the final instruction by combining the symptoms with a piece of text stating the disease should be in the differential diagnosis of patients with these symptoms
(Fig.~\ref{fig:fig1}).

\subsubsection{Generated Instructions}
\label{subsec:geninstructions}

While we collect instructions from trustworthy sources for the DDX datasets, we introduce scalable and controllable techniques to generate instructions that vary in their phrasing, granularity and style for evaluating model robustness to different sorts of instructions on the UCI datasets.

\paragraph{Method} To generate instructions (overview in Fig.~\ref{fig:annotation-overview}), we initially fit a simple model, such as a shallow rule set, on the task's full training data (1).
Next, we serialize the model's logic into text using a \textit{template} (2).
While the templates capture useful logic for solving the task, they are highly structured.
We use GPT-3 (3) to revise the templates into natural language by prompting it to convert the templates into a concise paragraph~\cite{NEURIPS2020_1457c0d6}.
We carefully review the generated instructions to ensure they are faithful to the template.
Next, we discuss two models, \textit{rule sets} and \textit{prototypes classifiers}, we use to generate instructions.

\paragraph{Rule Sets} 

Humans often express their understanding through logical rules~\cite{informedMLRueden, pmlr-v80-xu18h,  hu-etal-2016-harnessing}.
Motivated by these findings, we use a rule sets classifier to create instruction templates.
Rule sets consist of independent decision rules for predicting a class (Figure~\ref{fig:annotation-overview}).
Each decision rule contains an independent \texttt{if-then} statement followed by a prediction and confidence value~\cite{Singh_imodels_a_python_2021}. 

\paragraph{Prototypes} Humans also represent knowledge using conceptual prototypes, representing the ``best'' examples of concepts~\cite{OSHERSON198135, chenthisthat}.
We use a \textit{prototypes} classifier as an instruction template.
The prototypes classifier computes centroids for each class and uses the nearest centroid's class as the prediction (Figure~\ref{fig:annotation-overview}).

\paragraph{Instructions} We generate $10$ prototypes and rulesets natural language instructions for the datasets in \benchmark.
To make the instructions more intelligible, we limit their complexity.
We set the depth of the rulesets to $1$.
We select the $10$ most important features for the prototypes classifier by computing their average information gain from XGBoost.

\label{sec:experiments}
\begin{figure*}[t]
    \centering
    \includegraphics[width=\linewidth]{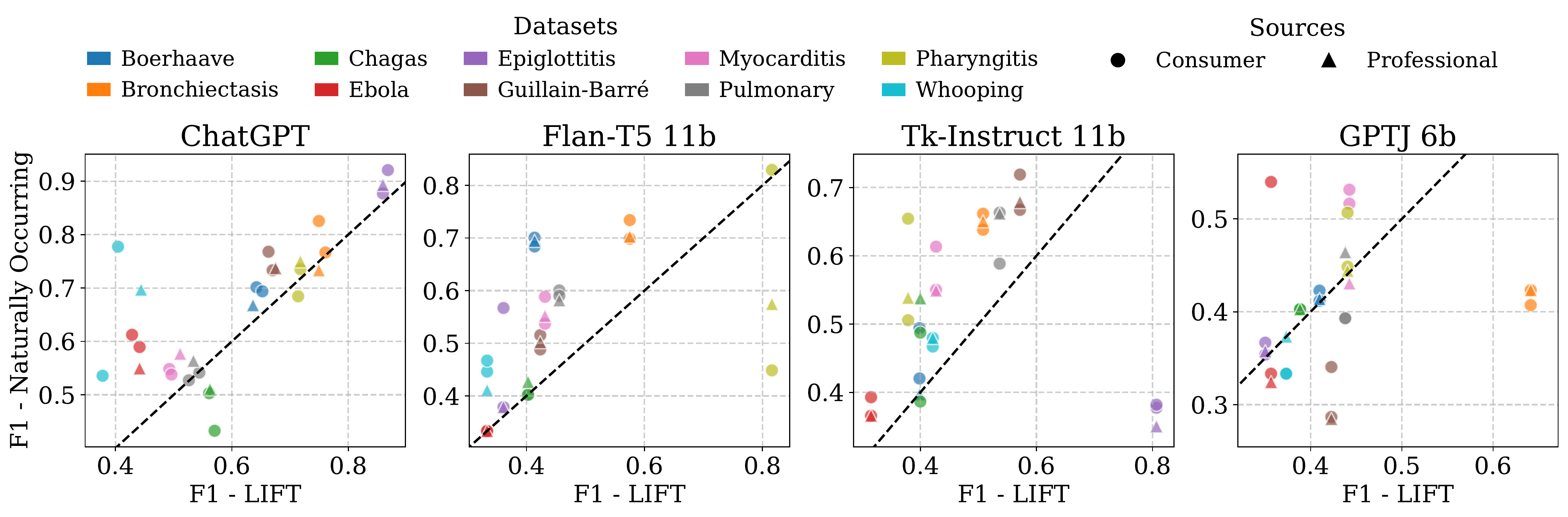}
    \caption{\textbf{Results on DDX tasks with Naturally Occurring Instructions.} Instructions greatly improve LLM generalization of ChatGPT, Flan-T5 11b, and Tk-Instruct 11b over the baseline without instructions (LIFT).}
    \label{fig:ddx-nat-lang-all-results}
\end{figure*}

\section{Experimental Setup}
In this section, we describe our experiment setup.

\paragraph{Models} We use pre-trained LLMs in our experiments~\cite{NEURIPS2020_1457c0d6}.
We employ two seq2seq models, Tk-Instruct 11b and Flan-T5 11b~\cite{supernaturalinstructions, flant5}.
These 11b T5 models received instruction fine-tuning, making them better at following instructions~\cite{2020t5}.
Also, we use GPT-J 6b, a causal language model, to compare a non-instruction tuned LLM~\cite{gpt-j}.
Last, we use ChatGPT, an LLM released by OpenAI trained with instruction tuning and reinforcement learning from human feedback (RLHF), a method that aligns LLMs with human preferences~\cite {chatgpt}.
Because OpenAI provides ChatGPT access through a costly API, we use this model in select experiments.

\paragraph{Prompts}
To construct prompts, we use the templates provided by Tk-Instruct and Flan-T5.
For example, Tk-Instruct prefers instructions to begin with \texttt{``Definition:''}.
Because GPT-J and ChatGPT do not have templates, we create prompts by joining the annotations from the \benchmark schema (subsection \ref{subsec:schema}) with line breaks. For GPT-J, we append \texttt{``The answer is''} so the next token will be the class.
For ChatGPT, we append \texttt{``Please answer with $c_1$,..,or $c_w$.''} because it is less amenable to direct completion.

\paragraph{Inference} For all models except ChatGPT, we set the temperature to 0 and take the generation from the prompted LLM as the predicted class.
However, ChatGPT often generates a paragraph of text instead of a label.
So, we use nucleus sampling fixing $p=0.1$.
We take the label that occurs in the generation to be the prediction.
However, label text may appear in other words, causing false positives (e.g., ``yes'' appears in ``eyes'').
So, we modified the class labels for ChatGPT to ``The answer is \texttt{$c_i$}'', which did not occur by chance in our evaluation.

\paragraph{Additional Baselines \& Metrics} 
To benchmark against an approach without instructions on \benchmark, we reproduce LIFT~\cite{lift}.
LIFT combines feature names, values, and a short description of the task into a prompt for in-context learning without instructions.
We create prompts for LIFT from \benchmark by omitting the instructions from the prompt.
When we compare LLMs with instructions against LIFT, we use the \textit{same} LLM, so the difference is the \textit{presence of instructions}.
Last, we evaluate using macro averaged F1 score for comparing across unbalanced labels.

\begin{figure*}[t]
    \centering
    \includegraphics[width=\linewidth]{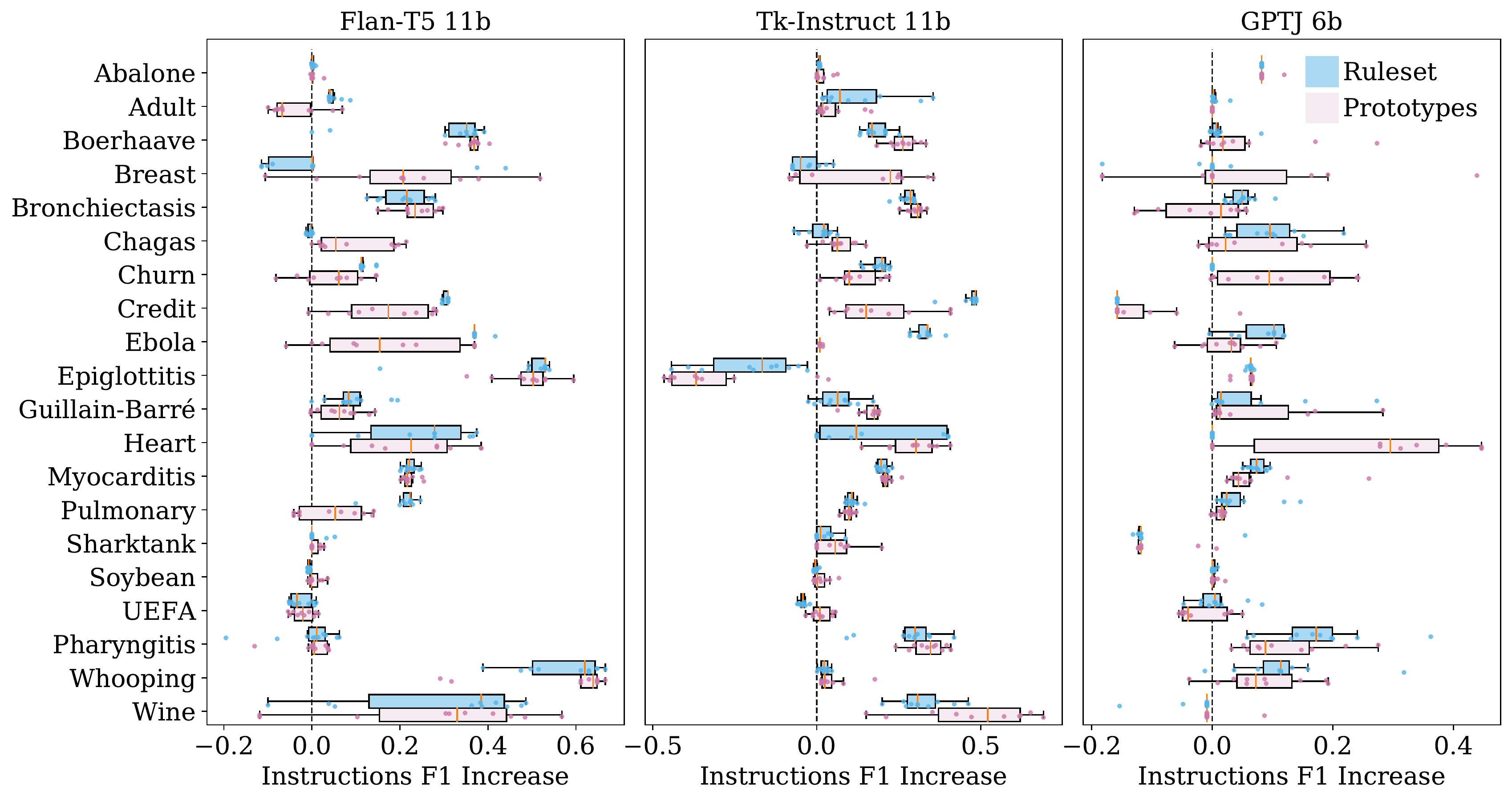}
    \caption{\textbf{Model results on all tasks with generated instructions.} Both prototypes and rulesets generated instructions lead to improved performance on most tasks for Flan-T5, Tk-Instruct, and GPT-J.}
    \label{fig:all-gen-results}
\end{figure*}

\section{Experiments}

\benchmark presents several new directions for evaluating how well instructions help LLMs use their knowledge for tabular prediction.
First, we evaluate in the \textit{zero-shot} setting and see whether LLMs perform well on tabular prediction with just instructions.
Next, we consider the \textit{few-shot} setting and assess LLM performance with both instructions and few-shot examples in-context.
Throughout, we use \benchmark to dissect the key limitations of current LLMs on tabular prediction tasks.

\subsection{Zero-Shot Performance}

Ideally, users could provide instructions to LLMs and achieve strong performance on tabular datasets without any data collection.
Therefore, we initially explore the zero-shot setting.

\paragraph{Performance} We compare the zero-shot F1 performance of LLMs using naturally occurring instructions with LIFT (no instructions) in Figure~\ref{fig:ddx-nat-lang-all-results}.
We use a Wilcoxon signed rank test to evaluate if instructions lead to significantly better performance and correct p-values using Holm–Bonferroni to account for multiple comparisons on the datasets~\cite{holm1979simple}.
The naturally occurring instructions significantly improve performance for ChatGPT, Flan-T5 11b, and Tk-Instruct 11b (p<0.01), while they are not helpful for GPT-J (p>0.05).
ChatGPT performs the best overall and scores quite highly for tasks such as Epiglottitis, reaching upwards of $0.9$ F1.
In Figure~\ref{fig:all-gen-results}, we compare LLM performance using generated instructions with LIFT.
We omit ChatGPT from these experiments due to costs.
Generated instructions significantly increase performance for all models (p<1e-4).
The improved GPT-J 6b performance with generated instructions is likely because the instructions include exact terms from the dataset, like the feature names, making them easier to follow.
Overall, instructions consistently benefit LLM performance on tabular prediction.

\paragraph{Differences Across Collection Sources} We assess differences in performance on the consumer and professional references.
For both Tk-Instruct 11b and Flan T5 11b, the instructions from the professional medical reference only performed best on $1/10$ data sets (Fig.~\ref{fig:ddx-nat-lang-all-results}).
However, the professional instructions performed best on $4/10$ datasets for ChatGPT.
These results could be explained by ChatGPT being a larger model capable of memorizing technical language during pre-training.
\begin{figure}[h!]
    \centering
    \includegraphics[width=.95\linewidth, trim={1cm 0.5cm 0.5cm 0.5cm}, clip]{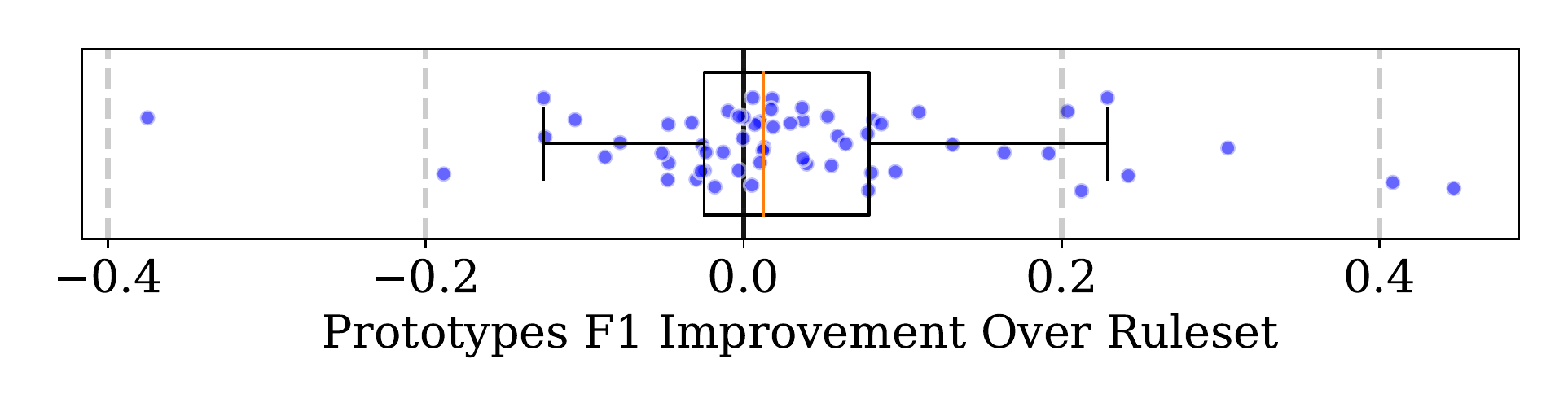}
    \caption{\textbf{Prototypes instructions perform better}.}
    \label{fig:comparing-rs-proto}
\end{figure}
\paragraph{Differences Across Generation Methods}  
We contrast performance on the rule sets and prototypes generated instructions using GPT-J 6b, Tk-Instruct 11b, and Flan-T5 11b.
Because the datasets have $10$ annotations for both types of instructions, we take the max performance on the instruction type to compare the upper bound performance.
The results provided in Figure~\ref{fig:comparing-rs-proto} show that the prototypes instruction type perform better ($p=0.03$).
This result indicates LLMs have an easier time interpreting conceptual prototypes than logical rules.

\paragraph{Generalization Beyond Instruction Logic}
Recall we derived the generated instructions from supervised models (Subsection~\ref{subsec:geninstructions}).
We evaluate whether LLMs with these instructions generalize better than the original supervised models.
Flan-T5 11b outperforms the prototypes classifiers by $33\%$ on average across all the datasets and the rulesets classifiers by $91\%$ (detailed results in Appendix Figure~\ref{fig:base-classifier-comparison}).
This result shows that LLMs leverage the instructions for generalization benefits beyond simply repeating the logic in the instruction.

\begin{figure}[t]
    \centering
    \includegraphics[width=.9\linewidth, trim={0 .5cm 1.5cm .5cm}, clip]{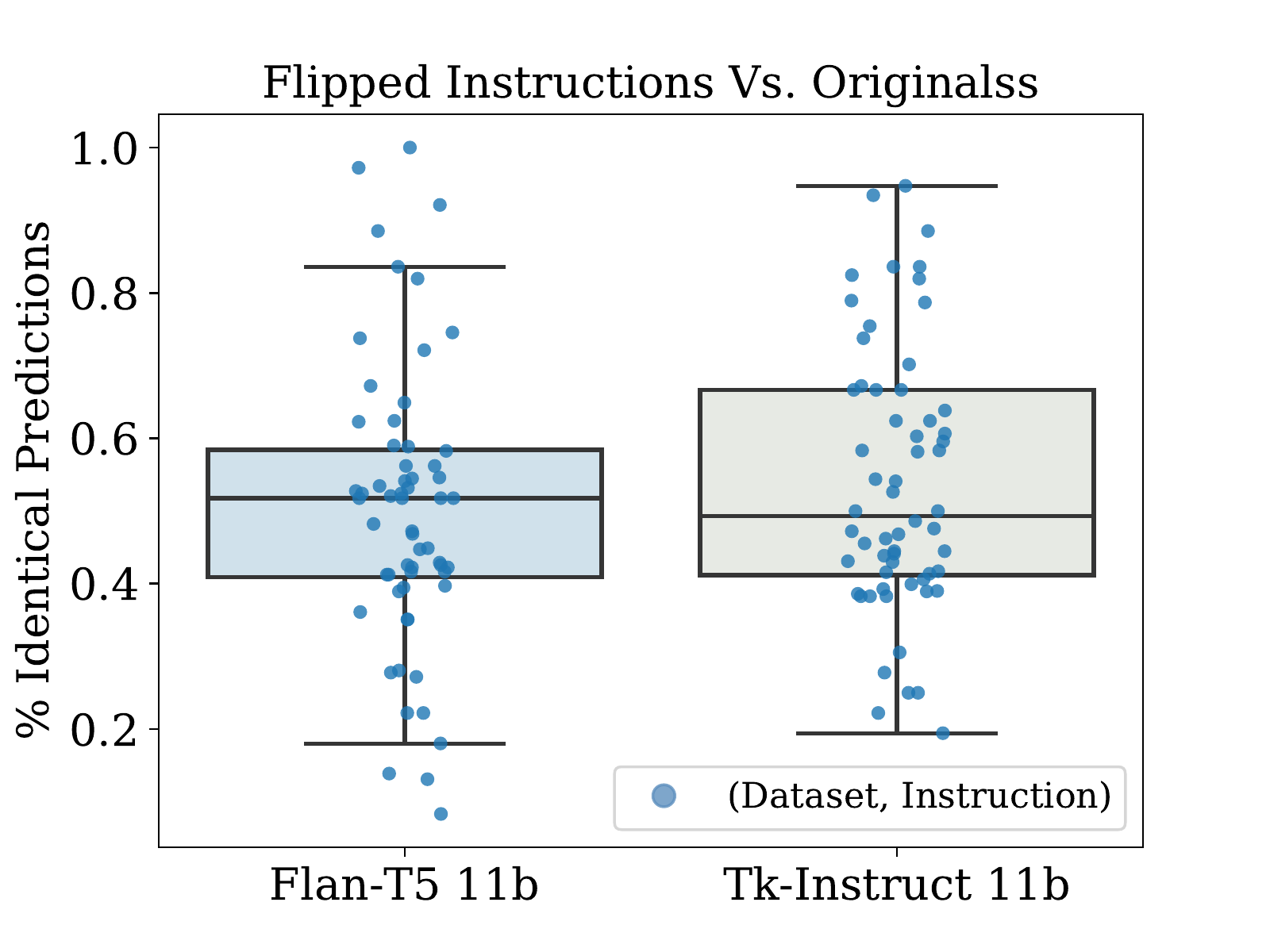}
    \caption{\textbf{LLMs predict many instances identically using instructions with \textit{flipped} logic}, indicating they are overly biased.}
    \label{fig:pct-instructions}
\end{figure}

\paragraph{Instruction Faithfulness}
We measure whether LLMs behave faithfully to instruction modifications.
We create a \textit{flipped} set of instructions by permuting the labels for the prototype class centroids and regenerating the instructions.
LLM predictions should be different following the flipped instructions because each centroid has a different class label.
We generate $10$ flipped instructions for the Boerhaave, Breast Cancer, Epiglottis, Heart Disease, Wine, and Bronchiectasis datasets.
We evaluate how often models predict the same class with the flipped instruction as the originals using Flan-T5 11b and Tk-Instruct 11b. We present the results in Figure~\ref{fig:pct-instructions}.
Surprisingly, the median number of identical predictions is $52\%$ for Flan and $49\%$ for Tk-Instruct.
This result demonstrates that LLMs may overly rely on their biases from pre-training instead of adapting to the in-context instructions.

\subsection{Few-Shot Performance}
\begin{figure}[t]
    \centering
    \includegraphics[width=.9\linewidth,trim={0 0 0 0},clip]{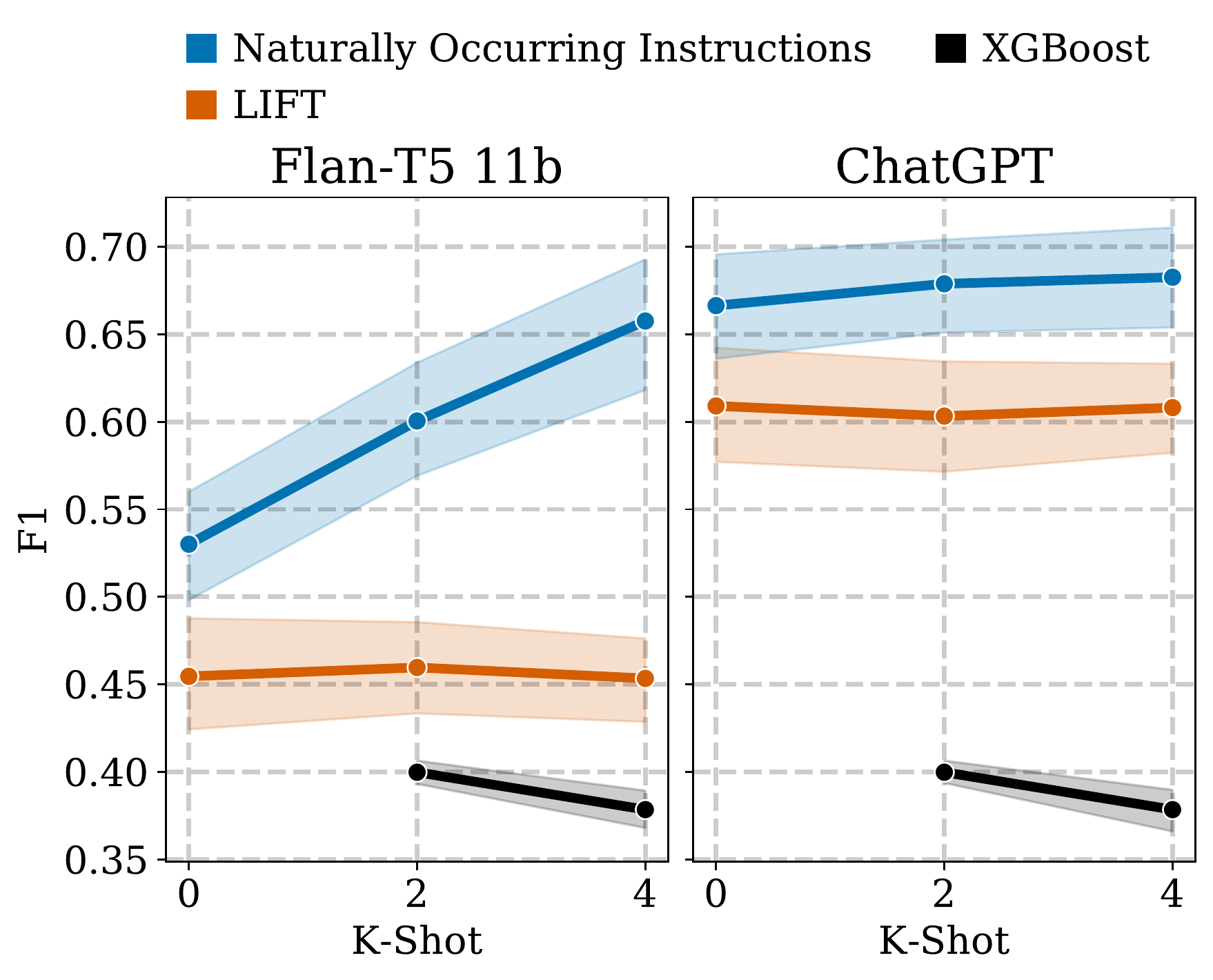}
    \caption{\textbf{Few-shot examples provide larger benefits to LLMs \textit{with} instructions than \textit{without} instructions.}}
    \label{fig:k-shot}
\end{figure}

Ideally, instructions alone would be sufficient to achieve good performance.
However, few-shot examples could significantly help models solve tasks, while still alleviating the challenge of gathering a large training set.
Therefore, this subsection studies using both instruction and few-shot examples for in-context learning.
For these experiments, we evaluate few-shot performance on the DDX tasks for $[0, 2, 4]$ shots due to Flan's context window, using both LIFT and the naturally occurring instructions.
We perform this evaluation with ChatGPT and Flan-T5 11b. To compare against state-of-the-art supervised models in this setting, we also evaluate few-shot XGBoost.
Additionally, because of the cost of ChatGPT, we evaluate a single seed.
Though, we perform few-shot sampling for $10$ datasets, which reduces any bias.
We present the results in Figure~\ref{fig:k-shot}.

\paragraph{Performance Gap} 

While zero-shot Flan-T5 with naturally occurring instructions outperforms LIFT by $20\%$, this gap increases to $44\%$ with $4$ in-context examples.
Similarly, ChatGPT's performance gap increases from $10\%$ to $13\%$ with $4$ examples. 
XGBoost scores only $0.38$ F1 with $4$-shot examples—much lower than the LLMs with instructions.
These results show that few-shot examples have compounding benefits for models with instructions, and LLMs significantly outperform supervised models in the few-shot setting.

\begin{figure}[t]
    \centering
    \includegraphics[width=\linewidth]{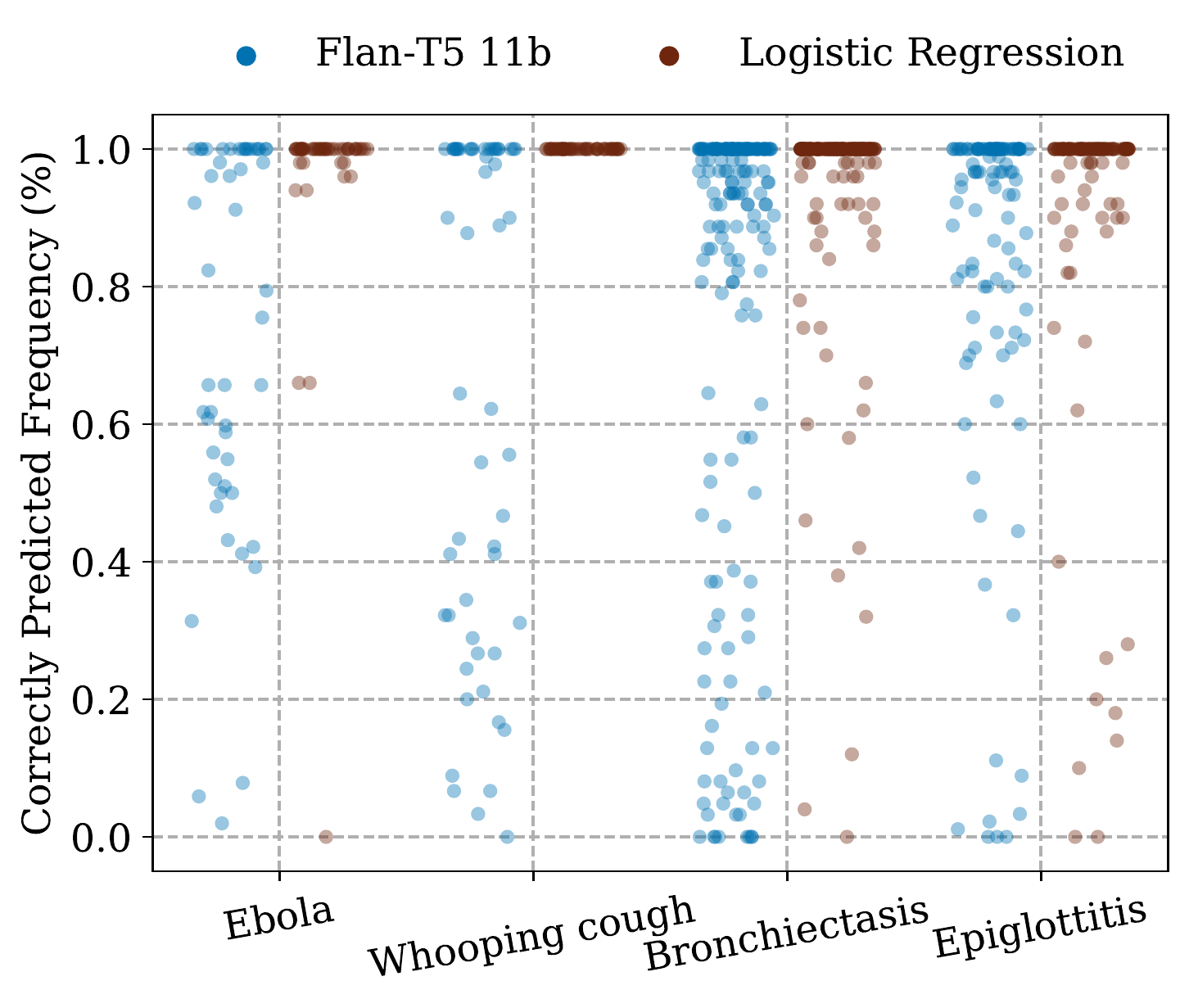}
    \caption{\textbf{LLMs + Instructions are highly biased against specific examples.} Each dot is a single data point. Over $30$ seeds, LLMs + Instructions with $4$-shot examples consistently misclassify specific data points.}
    \label{fig:bias-comp}
\end{figure}

\paragraph{Instance Learnability} Though the knowledge LLMs learn during pre-training helps performance, these models are overly biased and often fail to learn from few-shot examples.
We compute the percent of the time Flan-T5 11b, with naturally occurring instructions and $4$-shot examples, predicts each test datapoint correctly, over $30$ random seeds.
Also, we compare against logistic regression resampling the dataset with replacement $30$ times and perform this evaluation on the Ebola, Whooping cough, Bronchiectasis, and Epiglottitis datasets.
The results in Figure~\ref{fig:bias-comp} show Flan always mispredicts specific data points, indicating it is highly biased towards the opposite class label, and few-shot examples do little to help. 
Also, the logistic regression model struggles to learn fewer data points, indicating LLM biases is the leading cause of this result and not mislabeled or hard-to-learn examples.
Overall, the knowledge in LLMs is quite helpful for performance, but these models are often inflexible and overly biased.

\paragraph{Comparison to Fully Supervised} 
To evaluate how close LLMs are to outperforming fully supervised models, we compare Flan-T5 11b and ChatGPT with $4$-shot examples, against XGBoost fit on the entire training data.
XGBoost with all the data scores $0.94$ F1 on average on the DDX tasks, while ChatGPT $4$-shot scores on average $0.68$ and Flan-T5 11b scores $0.66$ F1.
Thus, there is still considerable room for improvement on \benchmark.

\paragraph{Few-Shot Selection}
We evaluate whether LLMs match fully supervised performance by selecting better few-shot examples instead of randomly sampling.
Though this experiment assume access to the complete training set, which is not our intended use case, it provides insight into whether approximate upper bound LLM performance matches fully supervised.
For each test point in the DDX tasks, we select $4$ few-shot examples using feature-weighted K-Nearest Neighbors (KNN) on the original data.
We compute feature weights using the average information gain of each feature from XGBoost fit on the training set.
This KNN model scores $0.93$ F1 on average on the tasks.
However, Flan T5 11b scores $0.77$ F1 using the same examples and naturally occurring instructions.
Even in this ideal setting, Flan-T5 still underperforms fully supervised.

\section{Related Work}
\label{sec:relatedwork}
In this section, we discuss related works.

\subsection{LLMs For Tabular Data}
Several prior works use LLMs for tabular data-related tasks.
Multiple works demonstrate LLMs are highly effective for tabular data augmentation tasks, such as generating new data or cleaning data~\cite{narayanAvanikaWrangle, borisov2023language, harari-katz-2022-shot}.
In addition, prior research show that similar architectures, such as transformer blocks, are effective for fine-tuning tabular data from scratch~\cite{huang2020tabtransformer, Gorishniy2021RevisitingDL}.
We focus on tabular prediction in sample efficient settings, leveraging the knowledge in pre-trained LLMs to improve performance.

Closer to our techniques, a few works show LLMs improve sample efficiency in the few-shot setting by serializing feature names and values into the prompt~\cite{lift, tabllm}.
These works consider fine-tuning and in-context learning approaches and achieve effective results.
Still, these works do not consider using instructions to improve sample efficiency nor introduce datasets with instruction annotations.

\subsection{Learning From Instructions}

Our work extends related research that shows instructions help increase the performance of LLMs on downstream NLP tasks~\cite{sanh2022multitask, supernaturalinstructions, min-etal-2022-metaicl, zhong-etal-2021-adapting-language}.
For instance, several works show that including task instructions during fine-tuning improves performance learning from instructions for different NLP tasks, such as summarization, sentiment analysis, and question answering \cite{wei2022finetuned, mishra-etal-2022-cross, ye2023guess}.
Others extend these findings to text embedding tasks and find instructions lead to improved downstream performance~\cite{Su2022OneEA}. 
While prior works consider improving performance on NLP or text-embedding tasks using LLMs tuned with instructions, we focus on answering whether instructions help LLMs utilize their world knowledge to improve performance on tabular data tasks.

\section{Discussion \& Conclusion}

In this work, we introduced \benchmark, a new benchmark of tabular datasets annotated with instructions, where the goal is to use the instructions to solve the prediction problem.
To support a wide range of possible use cases, we annotated the tasks in \benchmark with instructions that vary along several factors, including their collection source, granularity, and phrasing.
We presented results using several pre-trained LLMs, such as Flan-T5 11b and ChatGPT and found that instructions significantly increased task performance.

Improved methods for learning from the instructions for tabular prediction will significantly impact many domains.
Indeed, tabular data is the most common data format for many fields that rely on relational databases, such as medicine, finance, manufacturing, and climate science~\cite{shwartz-ziv2021tabular}.
Thus, learning from instructions has the potential to reduce the need for costly and time-consuming data collection procedures in many areas, offering widespread benefits.

The \benchmark benchmark provides the basis for future research into solving tabular prediction problems with instructions. 
So far, we have shown that instructions lead to improved performance overall and quite strong performance on specific tasks in the benchmark.
Still, several limitations of current LLMs hinder their performance, such as biases against certain instances and failures following instructions.
Accordingly, a promising direction for future research is developing models that strike a better balance between utilizing knowledge from pre-training when it is helpful to the task and faithfully adapting to new information.
Finally, because different application areas require domain-specific knowledge that LLMs may not have seen in their pre-training, another promising direction for future research is additional self-supervised training on domain-specific texts, e.g., medical references or manufacturing manuals.

\bibliography{anthology,custom}

\begin{thebibliography}{39}
\expandafter\ifx\csname natexlab\endcsname\relax\def\natexlab#1{#1}\fi

\bibitem[{Borisov et~al.(2023)Borisov, Sessler, Leemann, Pawelczyk, and
  Kasneci}]{borisov2023language}
Vadim Borisov, Kathrin Sessler, Tobias Leemann, Martin Pawelczyk, and Gjergji
  Kasneci. 2023.
\newblock \href {https://openreview.net/forum?id=cEygmQNOeI} {Language models
  are realistic tabular data generators}.
\newblock In \emph{The Eleventh International Conference on Learning
  Representations}.

\bibitem[{Brown et~al.(2020)Brown, Mann, Ryder, Subbiah, Kaplan, Dhariwal,
  Neelakantan, Shyam, Sastry, Askell, Agarwal, Herbert-Voss, Krueger, Henighan,
  Child, Ramesh, Ziegler, Wu, Winter, Hesse, Chen, Sigler, Litwin, Gray, Chess,
  Clark, Berner, McCandlish, Radford, Sutskever, and
  Amodei}]{NEURIPS2020_1457c0d6}
Tom Brown, Benjamin Mann, Nick Ryder, Melanie Subbiah, Jared~D Kaplan, Prafulla
  Dhariwal, Arvind Neelakantan, Pranav Shyam, Girish Sastry, Amanda Askell,
  Sandhini Agarwal, Ariel Herbert-Voss, Gretchen Krueger, Tom Henighan, Rewon
  Child, Aditya Ramesh, Daniel Ziegler, Jeffrey Wu, Clemens Winter, Chris
  Hesse, Mark Chen, Eric Sigler, Mateusz Litwin, Scott Gray, Benjamin Chess,
  Jack Clark, Christopher Berner, Sam McCandlish, Alec Radford, Ilya Sutskever,
  and Dario Amodei. 2020.
\newblock \href
  {https://proceedings.neurips.cc/paper/2020/file/1457c0d6bfcb4967418bfb8ac142f64a-Paper.pdf}
  {Language models are few-shot learners}.
\newblock In \emph{Advances in Neural Information Processing Systems},
  volume~33, pages 1877--1901. Curran Associates, Inc.

\bibitem[{Bullers(2016)}]{Bullers2016}
K.~Bullers. 2016.
\newblock \href {https://doi.org/10.3163/1536-5050.104.4.028} {Merck manuals}.
\newblock \emph{Journal of the Medical Library Association : JMLA},
  104(4):369--371.

\bibitem[{Cath(2018)}]{Cath2018-aj}
Corinne Cath. 2018.
\newblock Governing artificial intelligence: ethical, legal and technical
  opportunities and challenges.
\newblock \emph{Philos Trans A Math Phys Eng Sci}, 376(2133).

\bibitem[{Chen et~al.(2019)Chen, Li, Tao, Barnett, Su, and
  Rudin}]{chenthisthat}
Chaofan Chen, Oscar Li, Chaofan Tao, Alina~Jade Barnett, Jonathan Su, and
  Cynthia Rudin. 2019.
\newblock \emph{This Looks like That: Deep Learning for Interpretable Image
  Recognition}. Curran Associates Inc., Red Hook, NY, USA.

\bibitem[{Chung et~al.(2022)Chung, Hou, Longpre, Zoph, Tay, Fedus, Li, Wang,
  Dehghani, Brahma, Webson, Gu, Dai, Suzgun, Chen, Chowdhery, Narang, Mishra,
  Yu, Zhao, Huang, Dai, Yu, Petrov, Chi, Dean, Devlin, Roberts, Zhou, Le, and
  Wei}]{flant5}
Hyung~Won Chung, Le~Hou, Shayne Longpre, Barret Zoph, Yi~Tay, William Fedus,
  Eric Li, Xuezhi Wang, Mostafa Dehghani, Siddhartha Brahma, Albert Webson,
  Shixiang~Shane Gu, Zhuyun Dai, Mirac Suzgun, Xinyun Chen, Aakanksha
  Chowdhery, Sharan Narang, Gaurav Mishra, Adams Yu, Vincent Zhao, Yanping
  Huang, Andrew Dai, Hongkun Yu, Slav Petrov, Ed~H. Chi, Jeff Dean, Jacob
  Devlin, Adam Roberts, Denny Zhou, Quoc~V. Le, and Jason Wei. 2022.
\newblock \href {https://doi.org/10.48550/ARXIV.2210.11416} {Scaling
  instruction-finetuned language models}.

\bibitem[{Crawford and Schultz(2014)}]{crawford2014big}
Kate Crawford and Jason Schultz. 2014.
\newblock \href
  {https://www.microsoft.com/en-us/research/publication/big-data-and-due-process-toward-a-framework-to-redress-predictive-privacy-harms/}
  {Big data and due process: Toward a framework to redress predictive privacy
  harms}.
\newblock \emph{Boston College Law Review}, 55(1).

\bibitem[{Dinh et~al.(2022)Dinh, Zeng, Zhang, Lin, Gira, Rajput, yong Sohn,
  Papailiopoulos, and Lee}]{lift}
Tuan Dinh, Yuchen Zeng, Ruisu Zhang, Ziqian Lin, Michael Gira, Shashank Rajput,
  Jy~yong Sohn, Dimitris Papailiopoulos, and Kangwook Lee. 2022.
\newblock Lift: Language-interfaced fine-tuning for non-language machine
  learning tasks.
\newblock \emph{NeurIPS}.

\bibitem[{Dua and Graff(2017)}]{Dua:2019}
Dheeru Dua and Casey Graff. 2017.
\newblock \href {http://archive.ics.uci.edu/ml} {{UCI} machine learning
  repository}.

\bibitem[{Gorishniy et~al.(2021)Gorishniy, Rubachev, Khrulkov, and
  Babenko}]{Gorishniy2021RevisitingDL}
Yu.~V. Gorishniy, Ivan Rubachev, Valentin Khrulkov, and Artem Babenko. 2021.
\newblock Revisiting deep learning models for tabular data.
\newblock \emph{ArXiv}, abs/2106.11959.

\bibitem[{Harari and Katz(2022)}]{harari-katz-2022-shot}
Asaf Harari and Gilad Katz. 2022.
\newblock \href {https://doi.org/10.18653/v1/2022.acl-long.111} {Few-shot
  tabular data enrichment using fine-tuned transformer architectures}.
\newblock In \emph{Proceedings of the 60th Annual Meeting of the Association
  for Computational Linguistics (Volume 1: Long Papers)}, pages 1577--1591,
  Dublin, Ireland. Association for Computational Linguistics.

\bibitem[{Hegselmann et~al.(2022)Hegselmann, Buendia, Lang, Agrawal, Jiang, and
  Sontag}]{tabllm}
Stefan Hegselmann, Alejandro Buendia, Hunter Lang, Monica Agrawal, Xiaoyi
  Jiang, and David Sontag. 2022.
\newblock Tabllm: Few-shot classification of tabular data with large language
  models.
\newblock \emph{arXiv}.

\bibitem[{Holm(1979)}]{holm1979simple}
Sture Holm. 1979.
\newblock \href {http://www.jstor.org/stable/4615733} {A simple sequentially
  rejective multiple test procedure}.
\newblock \emph{Scandinavian Journal of Statistics}, 6(2):65--70.
\newblock Accessed 1 Apr. 2023.

\bibitem[{Hu et~al.(2016)Hu, Ma, Liu, Hovy, and Xing}]{hu-etal-2016-harnessing}
Zhiting Hu, Xuezhe Ma, Zhengzhong Liu, Eduard Hovy, and Eric Xing. 2016.
\newblock \href {https://doi.org/10.18653/v1/P16-1228} {Harnessing deep neural
  networks with logic rules}.
\newblock In \emph{Proceedings of the 54th Annual Meeting of the Association
  for Computational Linguistics (Volume 1: Long Papers)}, pages 2410--2420,
  Berlin, Germany. Association for Computational Linguistics.

\bibitem[{Huang et~al.(2020)Huang, Khetan, Cvitkovic, and
  Karnin}]{huang2020tabtransformer}
Xin Huang, Ashish Khetan, Milan Cvitkovic, and Zohar Karnin. 2020.
\newblock \href {http://arxiv.org/abs/2012.06678} {Tabtransformer: Tabular data
  modeling using contextual embeddings}.

\bibitem[{Hulsen(2020)}]{sharing-is-hard}
Tim Hulsen. 2020.
\newblock Sharing is {Caring-Data} sharing initiatives in healthcare.
\newblock \emph{Int J Environ Res Public Health}, 17(9).

\bibitem[{Min et~al.(2022)Min, Lewis, Zettlemoyer, and
  Hajishirzi}]{min-etal-2022-metaicl}
Sewon Min, Mike Lewis, Luke Zettlemoyer, and Hannaneh Hajishirzi. 2022.
\newblock \href {https://doi.org/10.18653/v1/2022.naacl-main.201} {{M}eta{ICL}:
  Learning to learn in context}.
\newblock In \emph{Proceedings of the 2022 Conference of the North American
  Chapter of the Association for Computational Linguistics: Human Language
  Technologies}, pages 2791--2809, Seattle, United States. Association for
  Computational Linguistics.

\bibitem[{Mishra et~al.(2022)Mishra, Khashabi, Baral, and
  Hajishirzi}]{mishra-etal-2022-cross}
Swaroop Mishra, Daniel Khashabi, Chitta Baral, and Hannaneh Hajishirzi. 2022.
\newblock \href {https://doi.org/10.18653/v1/2022.acl-long.244} {Cross-task
  generalization via natural language crowdsourcing instructions}.
\newblock In \emph{Proceedings of the 60th Annual Meeting of the Association
  for Computational Linguistics (Volume 1: Long Papers)}, pages 3470--3487,
  Dublin, Ireland. Association for Computational Linguistics.

\bibitem[{Moorthy and Ghosh(2015)}]{bigdataconsumerprivacymoothy}
Janakiraman Moorthy and Pulak Ghosh. 2015.
\newblock Big data and consumer privacy.
\newblock \emph{Vikalpa: The Journal for Decision Makers}, 40:92 -- 95.

\bibitem[{Murdoch(2021)}]{Murdoch2021}
Blake Murdoch. 2021.
\newblock \href {https://doi.org/10.1186/s12910-021-00687-3} {Privacy and
  artificial intelligence: challenges for protecting health information in a
  new era}.
\newblock \emph{BMC Medical Ethics}, 22(1):122.

\bibitem[{Narayan et~al.(2022)Narayan, Chami, Orr, and
  R\'{e}}]{narayanAvanikaWrangle}
Avanika Narayan, Ines Chami, Laurel Orr, and Christopher R\'{e}. 2022.
\newblock \href {https://doi.org/10.14778/3574245.3574258} {Can foundation
  models wrangle your data?}
\newblock \emph{Proc. VLDB Endow.}, 16(4):738–746.

\bibitem[{{National Library of Medicine (US)}()}]{medlineplus}
{National Library of Medicine (US)}.
\newblock \href {https://medlineplus.gov/} {Medlineplus}.
\newblock [updated Jun 24; cited 2020 Jul 1].

\bibitem[{Osherson and Smith(1981)}]{OSHERSON198135}
Daniel~N. Osherson and Edward~E. Smith. 1981.
\newblock \href {https://doi.org/https://doi.org/10.1016/0010-0277(81)90013-5}
  {On the adequacy of prototype theory as a theory of concepts}.
\newblock \emph{Cognition}, 9(1):35--58.

\bibitem[{Peng et~al.(2018)Peng, Tang, Lin, and Chang}]{NEURIPS2018_b5a1d925}
Yu-Shao Peng, Kai-Fu Tang, Hsuan-Tien Lin, and Edward Chang. 2018.
\newblock \href
  {https://proceedings.neurips.cc/paper/2018/file/b5a1d925221b37e2e399f7b319038ba0-Paper.pdf}
  {Refuel: Exploring sparse features in deep reinforcement learning for fast
  disease diagnosis}.
\newblock In \emph{Advances in Neural Information Processing Systems},
  volume~31. Curran Associates, Inc.

\bibitem[{Raffel et~al.(2020)Raffel, Shazeer, Roberts, Lee, Narang, Matena,
  Zhou, Li, and Liu}]{2020t5}
Colin Raffel, Noam Shazeer, Adam Roberts, Katherine Lee, Sharan Narang, Michael
  Matena, Yanqi Zhou, Wei Li, and Peter~J. Liu. 2020.
\newblock \href {http://jmlr.org/papers/v21/20-074.html} {Exploring the limits
  of transfer learning with a unified text-to-text transformer}.
\newblock \emph{Journal of Machine Learning Research}, 21(140):1--67.

\bibitem[{Sanh et~al.(2022)Sanh, Webson, Raffel, Bach, Sutawika, Alyafeai,
  Chaffin, Stiegler, Raja, Dey, Bari, Xu, Thakker, Sharma, Szczechla, Kim,
  Chhablani, Nayak, Datta, Chang, Jiang, Wang, Manica, Shen, Yong, Pandey,
  Bawden, Wang, Neeraj, Rozen, Sharma, Santilli, Fevry, Fries, Teehan, Scao,
  Biderman, Gao, Wolf, and Rush}]{sanh2022multitask}
Victor Sanh, Albert Webson, Colin Raffel, Stephen Bach, Lintang Sutawika, Zaid
  Alyafeai, Antoine Chaffin, Arnaud Stiegler, Arun Raja, Manan Dey, M~Saiful
  Bari, Canwen Xu, Urmish Thakker, Shanya~Sharma Sharma, Eliza Szczechla,
  Taewoon Kim, Gunjan Chhablani, Nihal Nayak, Debajyoti Datta, Jonathan Chang,
  Mike Tian-Jian Jiang, Han Wang, Matteo Manica, Sheng Shen, Zheng~Xin Yong,
  Harshit Pandey, Rachel Bawden, Thomas Wang, Trishala Neeraj, Jos Rozen,
  Abheesht Sharma, Andrea Santilli, Thibault Fevry, Jason~Alan Fries, Ryan
  Teehan, Teven~Le Scao, Stella Biderman, Leo Gao, Thomas Wolf, and Alexander~M
  Rush. 2022.
\newblock \href {https://openreview.net/forum?id=9Vrb9D0WI4} {Multitask
  prompted training enables zero-shot task generalization}.
\newblock In \emph{International Conference on Learning Representations}.

\bibitem[{Schulman et~al.(2022)Schulman, Zoph, Kim, Hilton, Menick, Weng,
  Uribe, Fedus, Metz, Pokorny, Lopes, Zhao, Vijayvergiya, Sigler, Perelman,
  Voss, Heaton, Parish, Cummings, Nayak, Balcom, Schnurr, Kaftan, Hallacy,
  Turley, Deutsch, Goel, Ward, Konstantinidis, Zaremba, Ouyang, Bogdonoff,
  Gross, Medina, Yoo, Lee, Lowe, Mossing, Huizinga, Jiang, Wainwright, Almeida,
  Lin, Zhang, Xiao, Slama, Bills, Gray, Leike, Pachocki, Tillet, Jain,
  Brockman, Ryder, Paino, Yuan, Winter, Wang, Bavarian, Babuschkin, Sidor,
  Kanitscheider, Pavlov, Plappert, Tezak, Jun, Zhuk, Pong, Kaiser, Tworek,
  Carr, Weng, Agarwal, Cobbe, Kosaraju, Power, Polu, Han, Puri, Jain, Chess,
  Gibson, Boiko, Parparita, Tootoonchian, Kosic, and Hesse}]{chatgpt}
John Schulman, Barret Zoph, Christina Kim, Jacob Hilton, Jacob Menick, Jiayi
  Weng, Juan Felipe~Ceron Uribe, Liam Fedus, Luke Metz, Michael Pokorny,
  Rapha~Gontijo Lopes, Shengjia Zhao, Arun Vijayvergiya, Eric Sigler, Adam
  Perelman, Chelsea Voss, Mike Heaton, Joel Parish, Dave Cummings, Rajeev
  Nayak, Valerie Balcom, David Schnurr, Tomer Kaftan, Chris Hallacy, Nicholas
  Turley, Noah Deutsch, Vik Goel, Jonathan Ward, Aris Konstantinidis, Wojciech
  Zaremba, Long Ouyang, Leonard Bogdonoff, Joshua Gross, David Medina, Sarah
  Yoo, Teddy Lee, Ryan Lowe, Dan Mossing, Joost Huizinga, Roger Jiang, Carroll
  Wainwright, Diogo Almeida, Steph Lin, Marvin Zhang, Kai Xiao, Katarina Slama,
  Steven Bills, Alex Gray, Jan Leike, Jakub Pachocki, Phil Tillet, Shantanu
  Jain, Greg Brockman, Nick Ryder, Alex Paino, Qiming Yuan, Clemens Winter, Ben
  Wang, Mo~Bavarian, Igor Babuschkin, Szymon Sidor, Ingmar Kanitscheider,
  Mikhail Pavlov, Matthias Plappert, Nik Tezak, Heewoo Jun, William Zhuk,
  Vitchyr Pong, Lukasz Kaiser, Jerry Tworek, Andrew Carr, Lilian Weng, Sandhini
  Agarwal, Karl Cobbe, Vineet Kosaraju, Alethea Power, Stanislas Polu, Jesse
  Han, Raul Puri, Shawn Jain, Benjamin Chess, Christian Gibson, Oleg Boiko, Emy
  Parparita, Amin Tootoonchian, Kyle Kosic, and Christopher Hesse. 2022.
\newblock \href {https://openai.com/blog/chatgpt} {Introducing chatgpt}.

\bibitem[{Shwartz-Ziv and Armon(2021)}]{shwartz-ziv2021tabular}
Ravid Shwartz-Ziv and Amitai Armon. 2021.
\newblock \href {https://openreview.net/forum?id=vdgtepS1pV} {Tabular data:
  Deep learning is not all you need}.
\newblock In \emph{8th ICML Workshop on Automated Machine Learning (AutoML)}.

\bibitem[{Singh et~al.(2021)Singh, Nasseri, Tan, Tang, and
  Yu}]{Singh_imodels_a_python_2021}
Chandan Singh, Keyan Nasseri, Yan~Shuo Tan, Tiffany Tang, and Bin Yu. 2021.
\newblock \href {https://doi.org/10.21105/joss.03192} {{imodels: a python
  package for fitting interpretable models}}.

\bibitem[{Su et~al.(2022)Su, Shi, Kasai, Wang, Hu, Ostendorf, tau Yih, Smith,
  Zettlemoyer, and Yu}]{Su2022OneEA}
Hongjin Su, Weijia Shi, Jungo Kasai, Yizhong Wang, Yushi Hu, Mari Ostendorf,
  Wen tau Yih, Noah~A. Smith, Luke Zettlemoyer, and Tao Yu. 2022.
\newblock One embedder, any task: Instruction-finetuned text embeddings.
\newblock \emph{ArXiv}, abs/2212.09741.

\bibitem[{Tchango et~al.(2022)Tchango, Goel, Wen, Martel, and
  Ghosn}]{tchango2022ddxplus}
Arsene~Fansi Tchango, Rishab Goel, Zhi Wen, Julien Martel, and Joumana Ghosn.
  2022.
\newblock \href {https://openreview.net/forum?id=heBKnuV42O} {{DDXP}lus: A new
  dataset for automatic medical diagnosis}.
\newblock In \emph{Thirty-sixth Conference on Neural Information Processing
  Systems Datasets and Benchmarks Track}.

\bibitem[{von Rueden et~al.(2023)von Rueden, Mayer, Beckh, Georgiev,
  Giesselbach, Heese, Kirsch, Pfrommer, Pick, Ramamurthy, Walczak, Garcke,
  Bauckhage, and Schuecker}]{informedMLRueden}
Laura von Rueden, Sebastian Mayer, Katharina Beckh, Bogdan Georgiev, Sven
  Giesselbach, Raoul Heese, Birgit Kirsch, Julius Pfrommer, Annika Pick,
  Rajkumar Ramamurthy, Michal Walczak, Jochen Garcke, Christian Bauckhage, and
  Jannis Schuecker. 2023.
\newblock \href {https://doi.org/10.1109/TKDE.2021.3079836} {Informed machine
  learning – a taxonomy and survey of integrating prior knowledge into
  learning systems}.
\newblock \emph{IEEE Transactions on Knowledge and Data Engineering},
  35(1):614--633.

\bibitem[{Wang and Komatsuzaki(2021)}]{gpt-j}
Ben Wang and Aran Komatsuzaki. 2021.
\newblock {GPT-J-6B: A 6 Billion Parameter Autoregressive Language Model}.
\newblock \url{https://github.com/kingoflolz/mesh-transformer-jax}.

\bibitem[{Wang et~al.(2022)Wang, Mishra, Alipoormolabashi, Kordi, Mirzaei,
  Arunkumar, Ashok, Dhanasekaran, Naik, Stap et~al.}]{supernaturalinstructions}
Yizhong Wang, Swaroop Mishra, Pegah Alipoormolabashi, Yeganeh Kordi, Amirreza
  Mirzaei, Anjana Arunkumar, Arjun Ashok, Arut~Selvan Dhanasekaran, Atharva
  Naik, David Stap, et~al. 2022.
\newblock Super-naturalinstructions:generalization via declarative instructions
  on 1600+ tasks.
\newblock In \emph{EMNLP}.

\bibitem[{Wei et~al.(2022)Wei, Bosma, Zhao, Guu, Yu, Lester, Du, Dai, and
  Le}]{wei2022finetuned}
Jason Wei, Maarten Bosma, Vincent Zhao, Kelvin Guu, Adams~Wei Yu, Brian Lester,
  Nan Du, Andrew~M. Dai, and Quoc~V Le. 2022.
\newblock \href {https://openreview.net/forum?id=gEZrGCozdqR} {Finetuned
  language models are zero-shot learners}.
\newblock In \emph{International Conference on Learning Representations}.

\bibitem[{{World Health Organization}(1993)}]{WHO1993}
{World Health Organization}. 1993.
\newblock \emph{The {ICD-10} Classification of Mental and Behavioural
  Disorders}.
\newblock World Health Organization.

\bibitem[{Xu et~al.(2018)Xu, Zhang, Friedman, Liang, and Van~den
  Broeck}]{pmlr-v80-xu18h}
Jingyi Xu, Zilu Zhang, Tal Friedman, Yitao Liang, and Guy Van~den Broeck. 2018.
\newblock \href {https://proceedings.mlr.press/v80/xu18h.html} {A semantic loss
  function for deep learning with symbolic knowledge}.
\newblock In \emph{Proceedings of the 35th International Conference on Machine
  Learning}, volume~80 of \emph{Proceedings of Machine Learning Research},
  pages 5502--5511. PMLR.

\bibitem[{Ye et~al.(2023)Ye, Kim, Jang, Shin, and Seo}]{ye2023guess}
Seonghyeon Ye, Doyoung Kim, Joel Jang, Joongbo Shin, and Minjoon Seo. 2023.
\newblock \href {https://openreview.net/forum?id=FtOxgKe_Zg2} {Guess the
  instruction! flipped learning makes language models stronger zero-shot
  learners}.
\newblock In \emph{The Eleventh International Conference on Learning
  Representations}.

\bibitem[{Zhong et~al.(2021)Zhong, Lee, Zhang, and
  Klein}]{zhong-etal-2021-adapting-language}
Ruiqi Zhong, Kristy Lee, Zheng Zhang, and Dan Klein. 2021.
\newblock \href {https://doi.org/10.18653/v1/2021.findings-emnlp.244} {Adapting
  language models for zero-shot learning by meta-tuning on dataset and prompt
  collections}.
\newblock In \emph{Findings of the Association for Computational Linguistics:
  EMNLP 2021}, pages 2856--2878, Punta Cana, Dominican Republic. Association
  for Computational Linguistics.

\end{thebibliography}
\bibliographystyle{acl_natbib}

\appendix

\clearpage
\pagebreak

\onecolumn
\section{Appendix}
\label{sec:appendix}

\subsection{Tasks}

In this appendix, we provide the tasks in \benchmark. There are $20$ tasks. $10$ tasks are differential diagnosis tasks (DDX)~\cite{tchango2022ddxplus}.
The rest are from the UCI repository~\cite{Dua:2019}.

\begin{table}[htbp]
\centering
\caption{UCI and DDX Datasets}
\label{tab:uci-ddx-datasets}
\begin{tabular}{ll}
\toprule
DDX & UCI \\
\midrule
Whooping Cough (A37) & Adult \\
Ebola (a98.4) &  Abalone \\
Chagas (B57) &  Breast Cancer \\
Guillain-Barre (G61.0) &  Churn \\
Pulmonary embolism (i26) & Heart Disease \\
Myocarditis (I51.4) &  Sharktank \\
Viral pharyngitis (J02.9) &  Soybean \\
Epiglottitis (J05.1) &  Credit \\
Bronchiectasis (J47) &  UEFA \\
Boerhaave (K22.3) &  Wine \\
\bottomrule
\end{tabular}
\end{table}

\subsection{Comparison To Template Models}

\begin{figure}[h]
    \centering
\includegraphics[width=.5\linewidth]{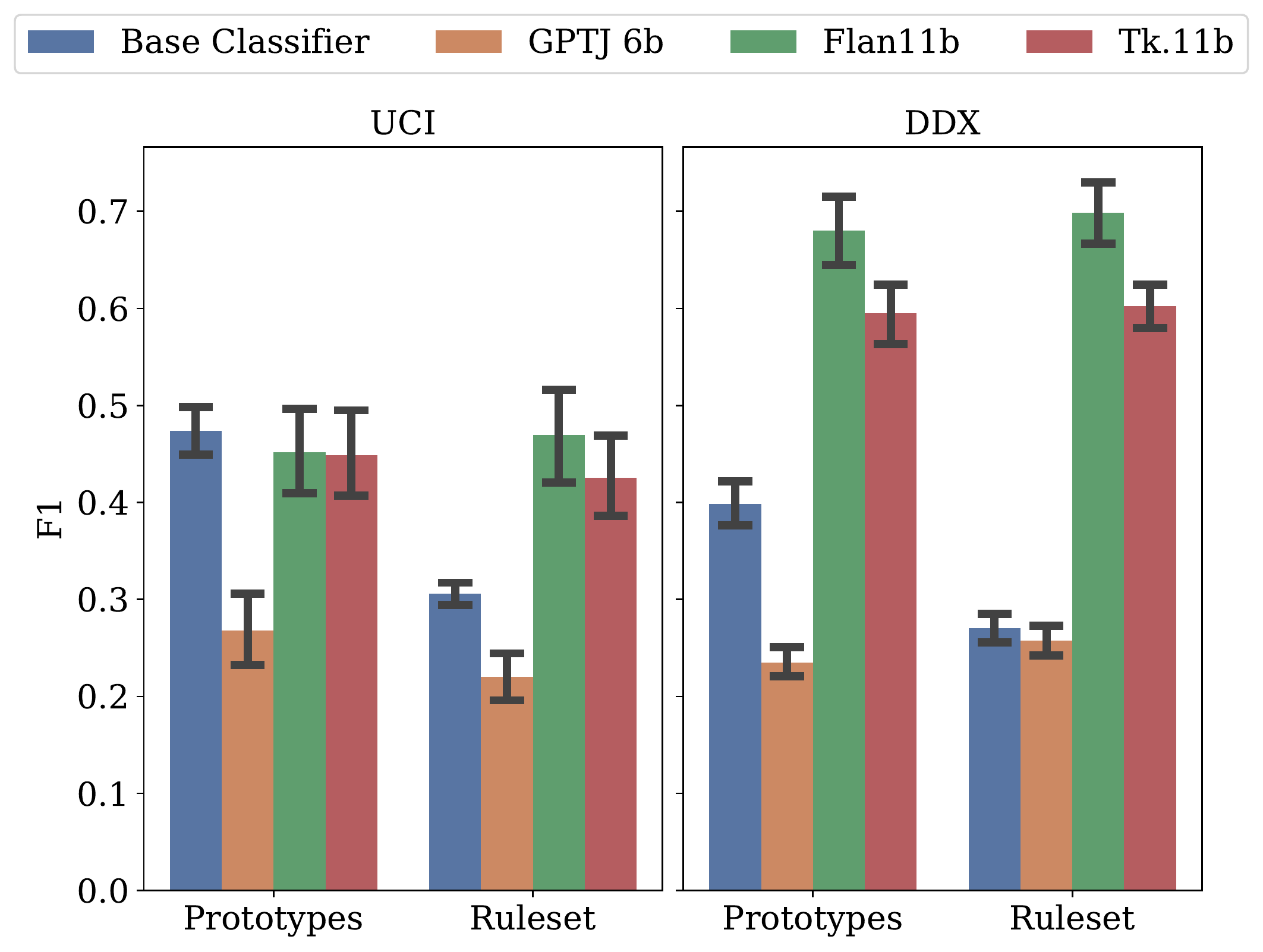}
    \caption{\textbf{LLMs with generated instructions outperform models used to create templates.} This result indicates generated instructions help align LLMs with task beyond the logic from the model.}
    \label{fig:base-classifier-comparison}
\end{figure}

In the main text, we show that LLMs with generated instructions outperform the ML models used to create the instruction templates.
In this appendix, we provide further details.
We show the performance of GPTJ 6b, Flan-T5 11b, and Tk Instruct 11b compared to the prototypes and rulesets classifiers in Figure~\ref{fig:base-classifier-comparison}.
We observe that both the Flan and Tk-Instruct models lead to considerable improvements over the models used to create the instruction templates except in the case of the prototypes classifier on the UCI dataset, where they perform about the same on average.
These results indicate that LLMs use the instructions to generalize beyond the simple models used to create the templates.
\subsection{Additional Few Shot Results Analysis}
\begin{figure*}[h]
    \centering
\includegraphics[width=.7\linewidth]{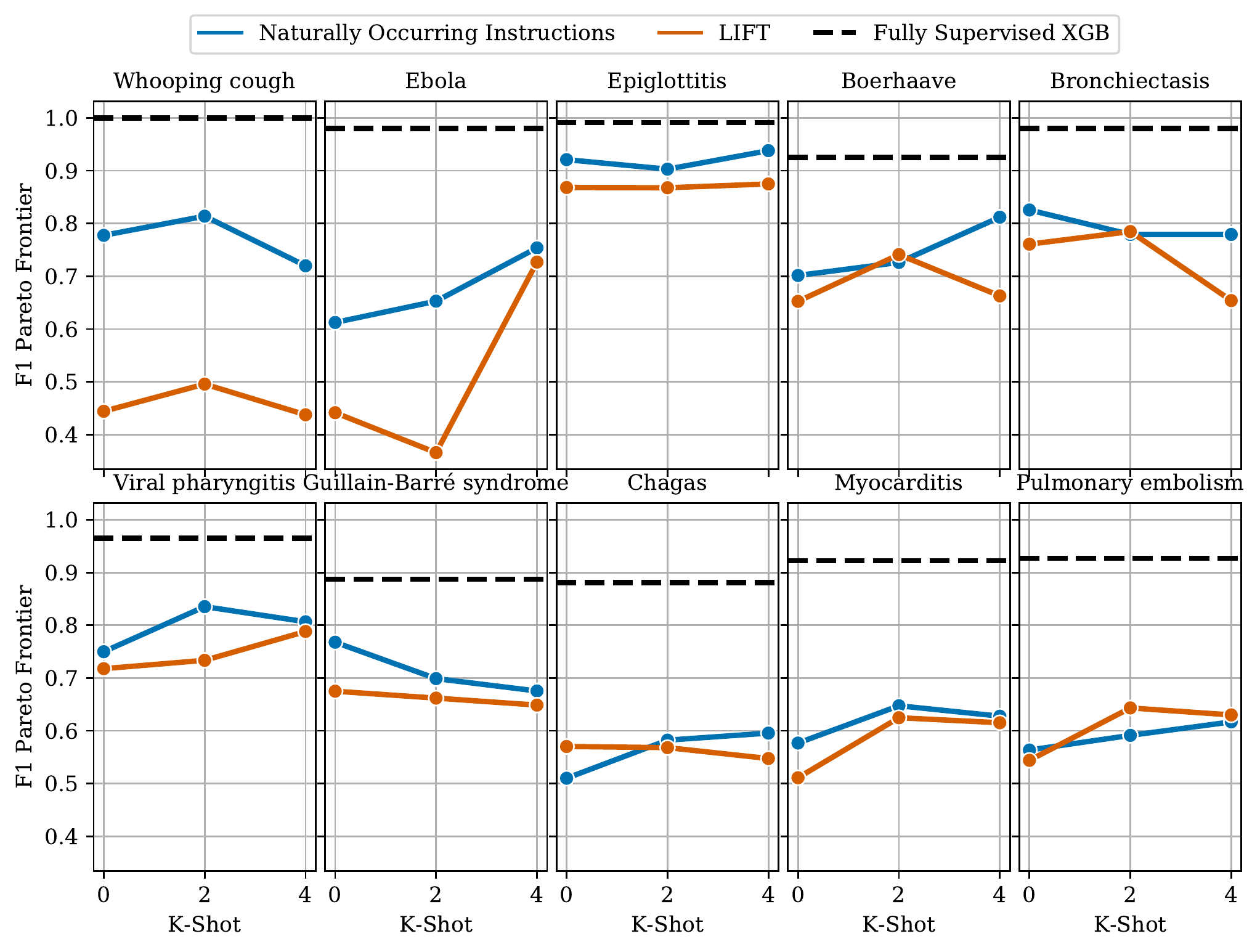}
    \caption{Pareto frontier across instructions for ChatGPT 0 - 4 shot}
    \label{fig:pareto-f1-chatgpt}
\end{figure*}

\begin{figure*}[h]
    \centering
\includegraphics[width=.7\linewidth]{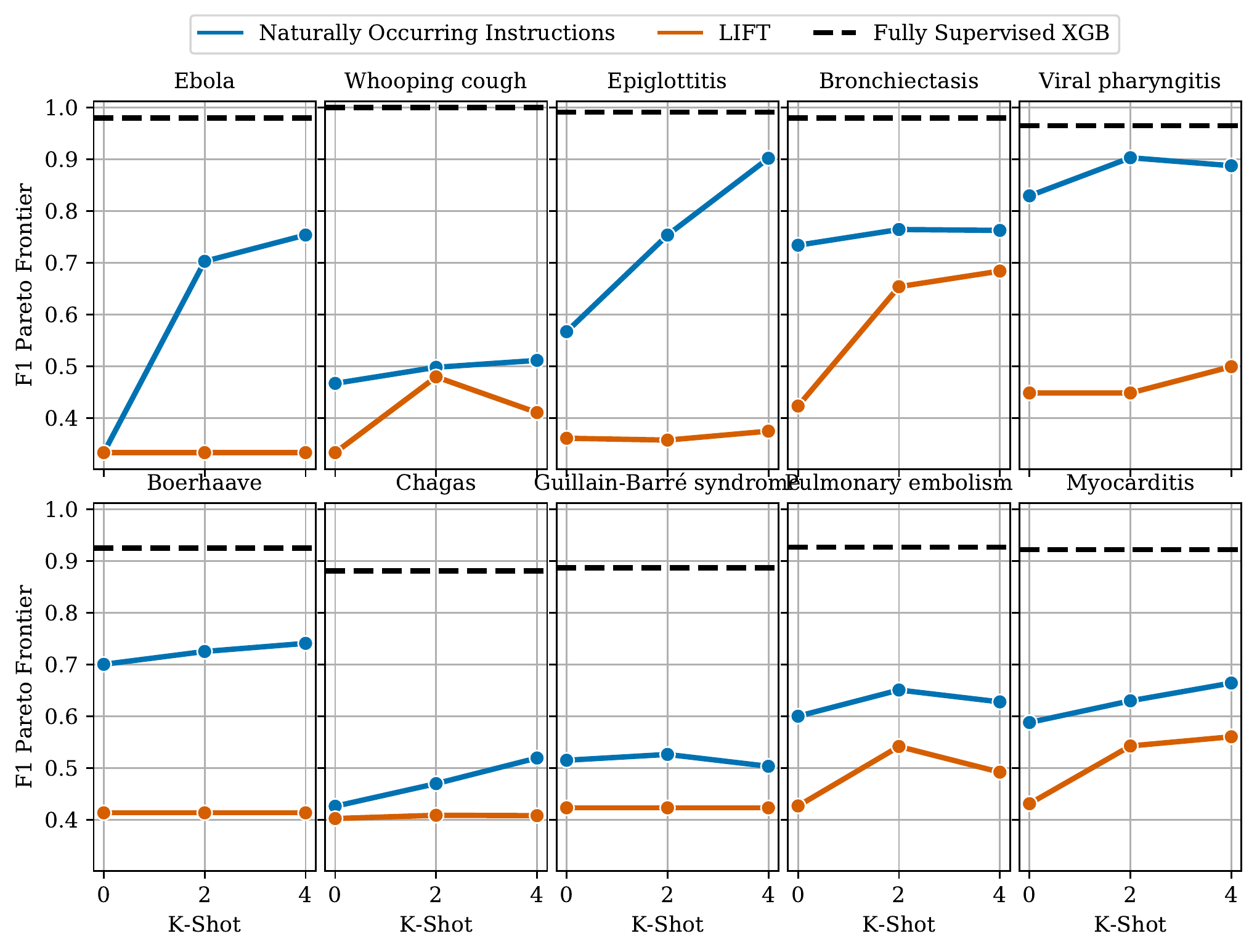}
    \caption{Pareto frontier across instructions for Flan 11b 0 - 4 shot}
    \label{fig:pareto-f1-flan}
\end{figure*}

In the main text, we presented aggregate results for the few shot performance of Flan-T5 11b and ChatGPT using the naturally occurring instructions in \benchmark.
In this appendix, we provide per dataset few-shot results.
We plot the pareto frontier of F1 score across each dataeset and shot, to help understand the performance of the best performing instructions.
We provide the results for ChatGPT in Figure~\ref{fig:pareto-f1-chatgpt} and the results for Flan-T5 11b in Figure~\ref{fig:pareto-f1-flan}.
For datasets like Epiglottitis, Boerhaave, and Guillain-Barre syndrome, the best performing models get quite close to fully supervised performance.
\clearpage
\subsection{Additional Zero Shot Results Analysis}
\begin{figure*}[h]
    \centering
\includegraphics[width=\linewidth]{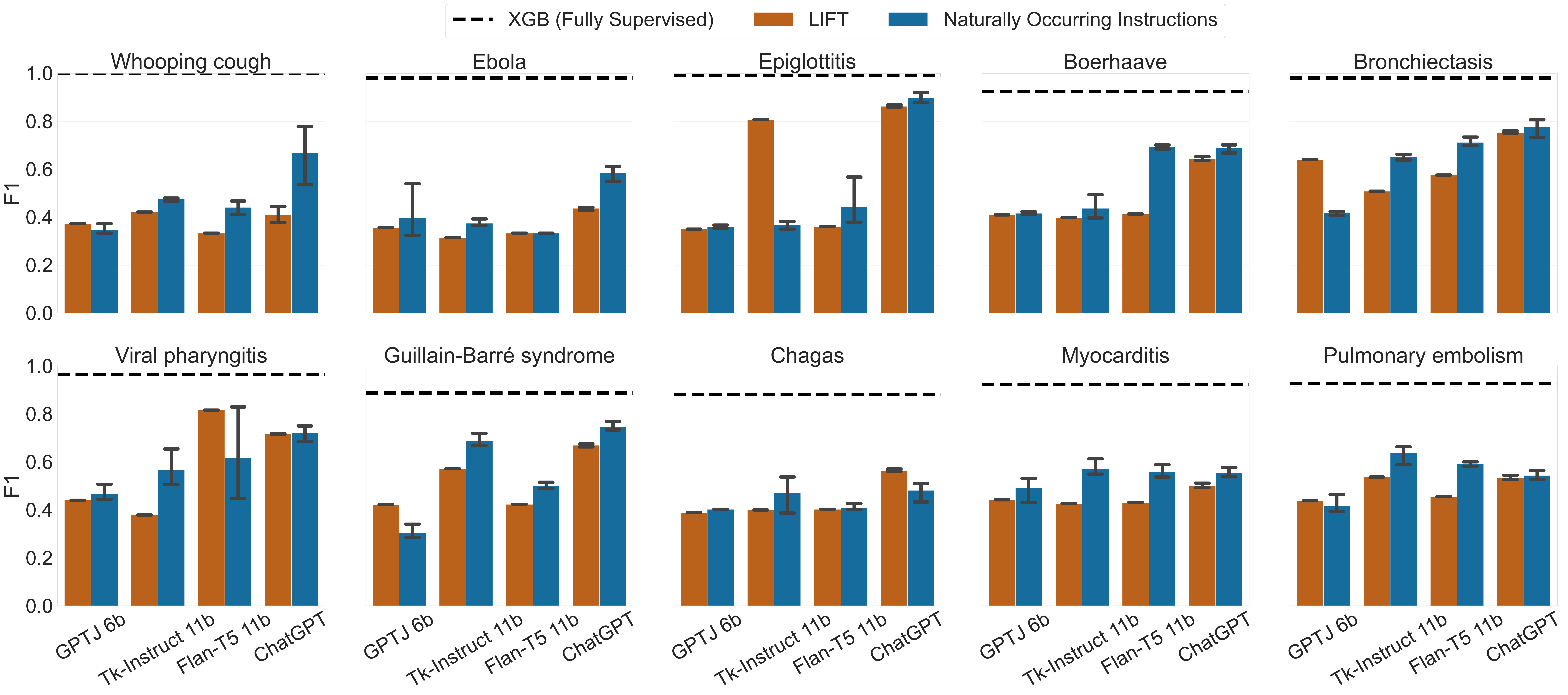}
    \caption{Overall results on ddx naturally occurring instructions.}
    \label{fig:ddx-nat-occur}
\end{figure*}

\begin{figure*}[h]
    \centering
    \includegraphics[width=\linewidth]{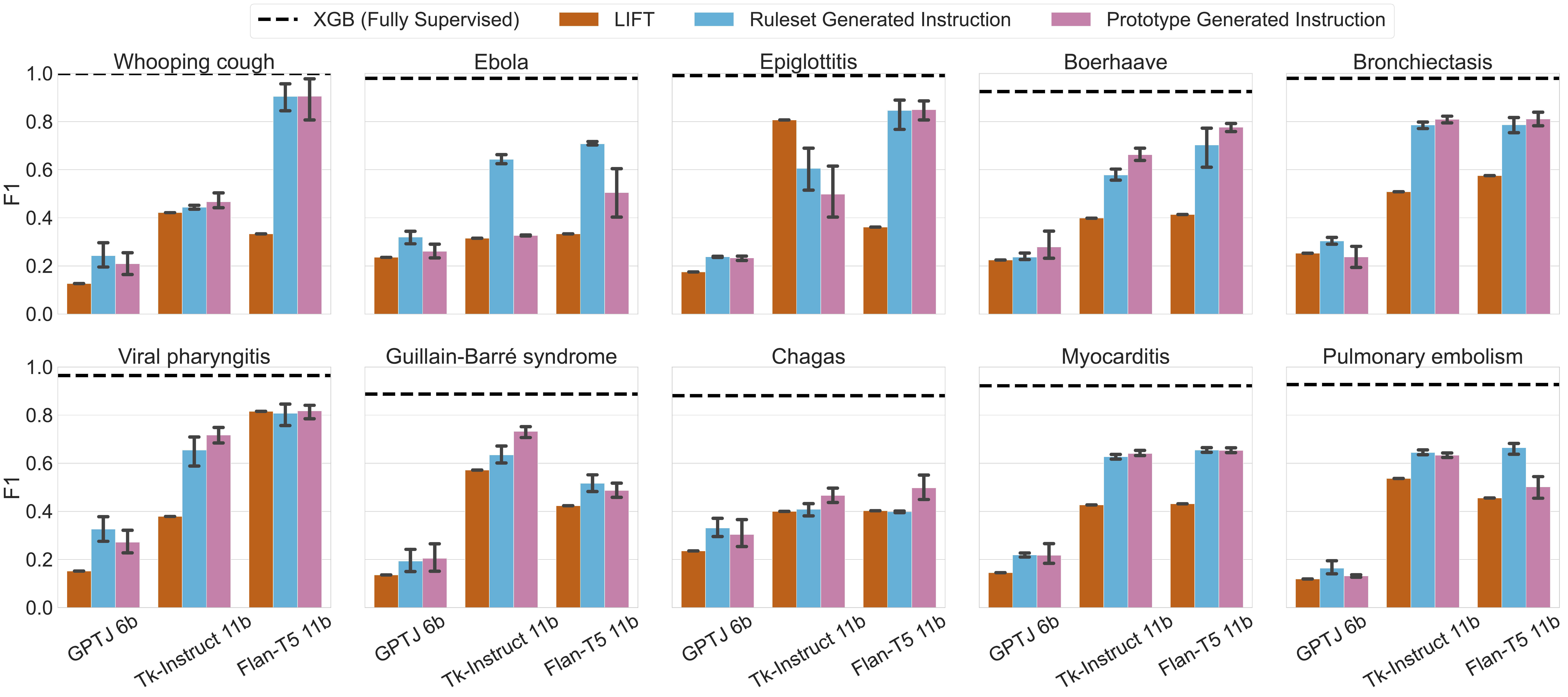}
    \caption{Overall results on ddx generated instructions.}
    \label{fig:gen-ddx-all}
\end{figure*}

\begin{figure*}[h]
    \centering
    \includegraphics[width=\linewidth]{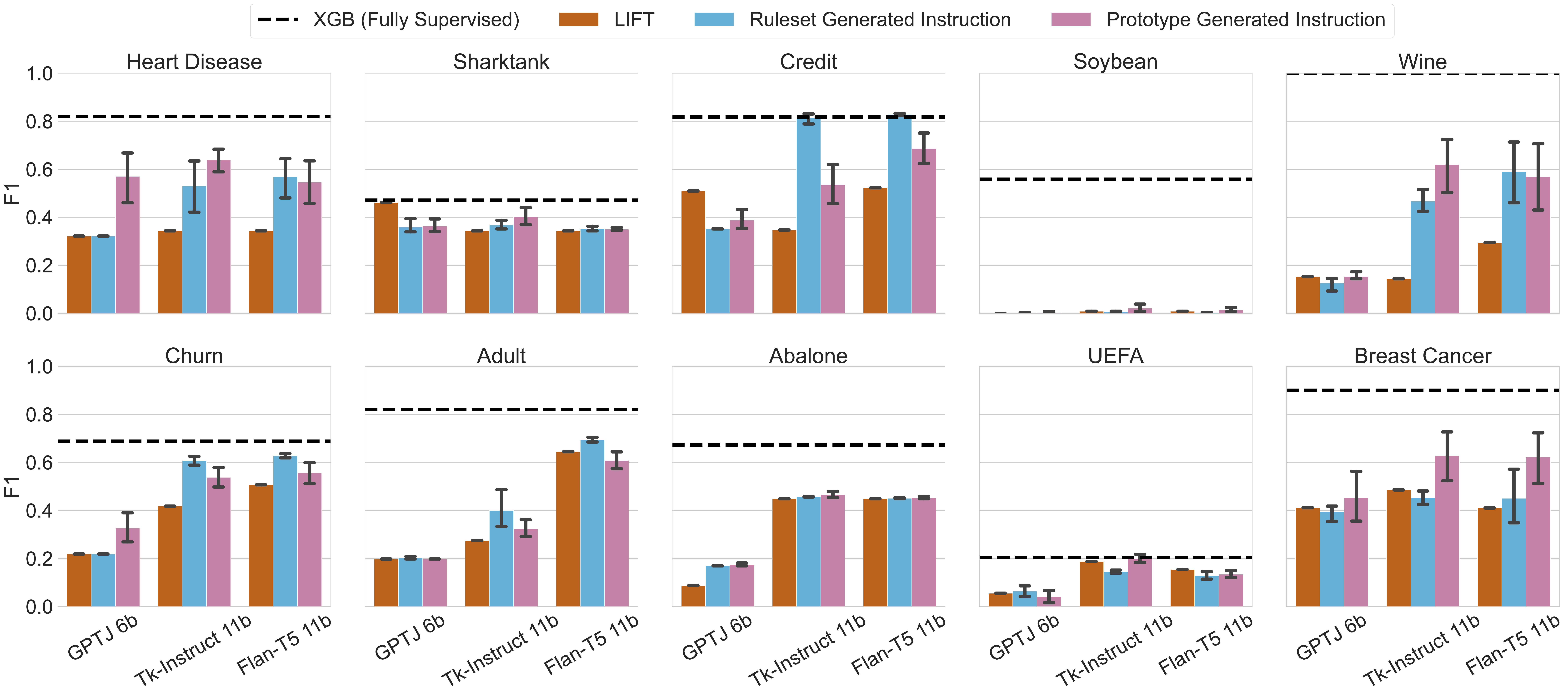}
    \caption{Overall results on uci generated instructions, including templates.}
    \label{fig:gen-uci-all}
\end{figure*}

In the main paper, we presented the zero shot performance of several models on datasets in \benchmark with the naturally occuring and generated instructions.
In this appendix, we provide these results broken out further by dataset.
The zero shot results for the naturally occurring instructions are provided in Figure~\ref{fig:ddx-nat-occur}.
The results for the generated instructions are given in Figure~\ref{fig:gen-ddx-all} and Figure~\ref{fig:gen-uci-all} for the DDX and UCI tasks respectively.
We additionally provide performance of a fully supervised XGB model for comparison.
We observe instructions lead to consistenly improved performance across the tasks.
\clearpage
\subsection{Instruction \& Prompt Examples}
In this appendix, we provide examples for the instructions and prompts in Tablet.
In Table~\ref{tab:proto-gen-examples}, we provide examples of the prototype generated instructions.
In Table~\ref{tab:rs-gen-examples}, we give examples of the rulesets generated instructions.
In Table~\ref{tab:nat-examples}, we include examples of the naturally occurring instructions.
 In Table~\ref{tab:ebola-prompt}, we provide an example from the Viral pharyngitis dataset.
This prompt includes a naturally occurring instruction collected from the NHS.
We can observe the patient exhibits many of the symptoms present in the instruction, such as a cough and throat pain.

\begin{table}[ht]
    \centering
    \begin{tabularx}{\textwidth}{lX}
       \toprule
       Dataset & Instruction \\ \midrule
        Churn & \texttt{Customers are more likely to churn if they have a month-to-month contract, fiber optic internet service, and phone service. Additionally, customers with shorter tenure (less than 8 months) and no additional services such as streaming movies, multiple lines, streaming tv, online backup, or tech support are also more likely to churn. Customers who have two year contracts, fiber optic internet service, and phone service, but do not have any of the additional services mentioned above, are less likely to churn.} \\  
        Ebola & \texttt{The answer to whether Ebola should be included in the differential diagnosis for a patient depends on their travel history. If the patient has not traveled out of the country in the past four weeks, then yes, it should be included in their differential diagnosis. Conversely, if the patient has recently traveled outside the country, then no, it does not need to be included as part of the diagnosis.} \\ Adult & \texttt{Generally, individuals who earn more than \$50K/yr tend to have higher levels of education (e.g., Bachelors or Prof-school), work in managerial or specialty occupations, and are married with a spouse. They also typically work longer hours per week (45+). On the other hand, those earning less than or equal to \$50K/yr usually have lower educational attainment (e.g., HS-grad) and work in clerical or craft-repair positions. Additionally, they tend to be unmarried or not in a family relationship and work fewer hours per week (less than 45).} \\ \bottomrule
    \end{tabularx}
    \caption{\textbf{Prototypes generated instruction examples}.}
    \label{tab:proto-gen-examples}
\end{table}

\begin{table}[ht]
    \centering
    \begin{tabularx}{\textwidth}{lX}
       \toprule
       Dataset & Instruction \\ \midrule
        Churn & \texttt{When predicting customer churn, if the contract type is month-to-month then the customer is likely to churn (74\% probability). For any other contract types, the customer is unlikely to churn (26\% probability).} \\ 
        Ebola & \texttt{If a patient has not traveled outside of their country in the last four weeks, there is a 50\% probability that Ebola should be included in the differential diagnosis. However, if they have travelled outside of the country, then it should definitely be included in the diagnosis with 100\% certainty.} \\ Adult & \texttt{The likelihood of a person having an income over \$50,000 per year significantly increases when they are married with a civil spouse, with a 76\% probability. Conversely, those who are not married with a civil spouse are likely to have an income of less than or equal to \$50,000 per year, with a 24\% probability.} \\ \bottomrule
    \end{tabularx}
    \caption{\textbf{Ruleset generated instruction examples}.}
    \label{tab:rs-gen-examples}
\end{table}

\begin{table}[ht]
    \centering
    \begin{tabularx}{\textwidth}{lX}
       \toprule
       Dataset & Instruction \\ \midrule
        Ebola & \texttt{Patients that have Ebola have symptoms that usually include:
- Fever
- Headache
- Joint and muscle aches
- Weakness and fatigue
- Sore throat
- Gastrointestinal symptoms including abdominal (belly) pain, diarrhea, and vomiting
- Loss of appetite
- Unexplained bleeding or bruising
- Other symptoms may include a rash, red eyes, and hiccups.
Overall, if the patient has conditions similar to those described above, the answer is yes, Ebola should be in the differential diagnosis. Otherwise, the answer is no.} \\ 
Bronchiectasis & \texttt{Patient's that have bronchiectasis have symptoms such as:
- A daily cough that occurs over months or years
- Daily production of large amounts of sputum (spit), which you cough up and which may have mucus, trapped particles, and pus
- Shortness of breath and wheezing (a whistling sound when you breathe)
- Chest pain
- Clubbing (the flesh under your fingernails and toenails gets thicker, causing the nails to curve downward)
...
- Chronic (long-term) pulmonary aspiration, which can inflame the airways
- Connective tissue diseases, such as rheumatoid arthritis, Sjögren’s syndromeexternal link, and Crohn’s diseaseexternal link
Overall, if the patient has large amounts of sputum, cough, and have conditions that damage their lungs the answer is yes, bronchiectasis should be in the differential diagnosis. Otherwise, the answer is no.
} \\
Viral pharyngitis & \texttt{Viral pharyngitis has the following symptoms: - a painful throat, especially when swallowing - a dry, scratchy throat - redness in the back of your mouth - bad breath - a mild cough - swollen neck glands The symptoms are similar for children, but children can also get a temperature and appear less active. If the patient has similar symptoms, then the answer is yes, Viral pharyngitis should be included in the differential diagnosis. Otherwise, the answer is no.} \\
        \bottomrule
    \end{tabularx}
    \caption{\textbf{Naturally Occurring instruction examples}.}
    \label{tab:nat-examples}
\end{table}

\begin{table}[ht]
    \centering
    \begin{tabularx}{\textwidth}{X}
       \toprule
       Viral pharyngitis dataset \\ \midrule
       Prompt: \texttt{Use the instructions to determine if Viral pharyngitis should be included in a differential diagnosis for the patient. Since there are a lot of different conditions that often share similar symptoms, a differential diagnosis lists the possible conditions that could cause the symptoms. Here are instructions for the differential diagnosis of Viral pharyngitis:}
       
\texttt{Viral pharyngitis has the following symptoms:}

\quad - \texttt{a painful throat, especially when swallowing}

\quad - \texttt{a dry, scratchy throat}

\quad - \texttt{redness in the back of your mouth}

\quad - \texttt{bad breath}

\quad - \texttt{a mild cough}

\quad - \texttt{swollen neck glands}

\texttt{The symptoms are similar for children, but children can also get a temperature and appear less active.
If the patient has similar symptoms, then the answer is yes, Viral pharyngitis should be included in the differential diagnosis. Otherwise, the answer is no. Answer with one of the following: no | yes.}

\texttt{Here are the patient's responses to questions about their symptoms.}

\texttt{Are you immunosuppressed?: yes}

\texttt{Characterize your pain: burning}

\texttt{Do you attend or work in a daycare?: yes}

\texttt{Do you feel pain somewhere?: palace, thyroid cartilage, right tonsil, and under the jaw}

\texttt{Do you have a cough?: yes}

\texttt{Do you have nasal congestion or a clear runny nose?: yes}

\texttt{Does the pain radiate to another location?: nowhere}

\texttt{Have you been in contact with a person with similar symptoms in the past 2 weeks?: yes}

\texttt{Have you traveled out of the country in the last 4 weeks?: no}

\texttt{How fast did the pain appear?: 5/10}

\texttt{How intense is the pain?: 7/10}

\texttt{How precisely is the pain located?: 10/10}

\texttt{Answer:}
 \\ \midrule
Label: \texttt{yes} \\      
        \bottomrule
    \end{tabularx}
    \caption{\textbf{An example of an entire prompt for an instances labeled \texttt{yes} from the Viral pharyngitis dataset.} This prompt uses a naturally occurring instruction sourced from the NHS and is formatted in the Flan prompt template.}
    \label{tab:ebola-prompt}
\end{table}

\end{document}